\documentclass[11pt]{article}

\usepackage[final]{acl}

\usepackage{times}
\usepackage{latexsym}
\usepackage{microtype}
\usepackage{graphicx}
\usepackage{xcolor}
\usepackage{url}
\usepackage{xspace} 
\usepackage{amsmath}
\usepackage{amssymb}
\usepackage{fix-cm}
\usepackage{mathtools}
\usepackage{amsthm}

\usepackage{booktabs}
\usepackage{multirow}
\usepackage{subcaption}
\usepackage{wrapfig}
\usepackage{siunitx}
\usepackage{tablefootnote}
\usepackage{tabularray}
\UseTblrLibrary{longtable} 

\usepackage{enumitem}
\usepackage{algpseudocode} 

\usepackage{hyperref}

\theoremstyle{plain}

\theoremstyle{definition}
\theoremstyle{remark}

\usepackage[textsize=tiny]{todonotes}
\newcommand{\mname}{{AlignDiff}\xspace} 
\usepackage[T1]{fontenc}
\usepackage[utf8]{inputenc}

\usepackage{microtype}

\usepackage{inconsolata}

\usepackage{graphicx}

\title{AlignDiff: Exploiting Model-Intrinsic Information for Better Preference Data Selection}

\author{%
    Peng Lai$^{1}$\thanks{Equal contribution.}, \
    He Zhu$^{2}$\protect\footnotemark[1], \
    Zhiwen Ruan$^{1}$, \
    Dongdong Zhang$^{3}$,\\
    \textbf{Yun Chen}$^{4}$,
    \textbf{Peng Li}$^{5}$,
    \textbf{Furu Wei}$^{3}$,
    \textbf{Yang Liu}$^{5}$,
    \textbf{Guanhua Chen}$^{1}\thanks{\ \ Corresponding author.}$ \\
    $^1$Southern University of Science and Technology, 
    $^2$Peking University, \\
    $^3$MSRA,
    $^4$Shanghai University of Finance and Economics,
    $^5$Tsinghua University  \\
}
\begin{document}
\maketitle
\begin{abstract}
Aligning large language models with human preferences remains a challenge, primarily due to the critical role of preference data quality in effective alignment. Existing datasets are frequently plagued by inherent noise and distribution shifts, which inherently limit model performance. To bridge this gap, we propose \mname, a preference data filtering framework driven by intrinsic model signals. \mname first identifies samples with clear preferences using both positive and inverse signals, then prioritizes the more challenging samples based on the average negative log-likelihood gap, encouraging the model to learn richer information from them. \mname is evaluated on two widely used model families (LLaMA and Qwen) and three benchmarks widely adopted in the alignment community (AlpacaEval 2.0, Arena-Hard, and MT-Bench). Across all settings, it consistently outperforms seven strong baselines. We conduct comprehensive ablation studies to validate the effectiveness of \mname, and further show that difficulty-based curriculum learning improves model performance. 
\end{abstract}

\section{Introduction}
As large language models (LLMs) advance, aligning outputs with human expectations and mitigating risks remains a major challenge~\citep{zhao2025surveylargelanguagemodels,jiang2024survey}. Early post-training efforts relied on preference alignment algorithms such as RLHF~\citep{rlhf} and DPO~\citep{dpo}, trained on preference data, to guide model behavior toward human values~\citep{wang2023aligninglargelanguagemodels}.
As research progresses, reinforcement fine-tuning (e.g., using the GRPO algorithm~\citep{shao2024deepseekmathpushinglimitsmathematical}) enhances model reasoning and alignment, potentially reducing reliance on traditional preference alignment. However, \citet{lanchantin2025bridgingofflineonlinereinforcement} found that semi-online DPO can match the performance of fully online GRPO while substantially decreasing training costs, indicating that further research on DPO is valuable. As the core of DPO, preference data provides models with more reliable preference signals and lays a solid foundation for subsequent method improvements~\citep{simpo,kto}. Early studies typically trained DPO directly on large-scale preference datasets such as UltraFeedback~\citep{ultrafeedback}, aiming to cover a broad human-preference space in one shot. However, follow-up work~\citep{xiao2025findingsweetspotpreference,shen2024datacentricrlhfsimplemetrics} has shown that simply “piling on data” introduces noise and distribution shifts that degrade alignment performance, highlighting the need to filter the data~\citep{wang2024secretsrlhflargelanguage} carefully.

To address this issue, previous work~\citep{f-dpo,hu2024comprehensivepreferencedatacollection} has proposed various preference data filtering methods that are grounded in external signals (e.g., scores produced by LLM-as-a-judge), leveraging external reward scores~\citep{yasunaga2024almaalignmentminimalannotation,curry-dpo} or data attributes~\citep{rip}. However, those methods may introduce preference data that does not align with the model’s intrinsic preferences (e.g., data with very low implicit reward margins), which can negatively affect the stability of alignment~\citep{xu2024dposuperiorppollm,wang2025incodpobalancingdistributionshift}. To mitigate these issues, recent studies have increasingly leveraged the implicit rewards of DPO for preference data filtering~\citep{gao2025principleddataselectionalignment,kim2025spreadpreferenceannotationdirect,chen2025bootstrappinglanguagemodelsdpo,xue2026reason,qiyuan2025efficient}. Nevertheless, such methods still exploit only a small portion of the model’s internal information, leaving many latent signals unexplored. This limitation not only constrains their effectiveness but also exacerbates the “squeezing effect” in DPO optimization: training on samples with the highest implicit reward margin tends to shrink the high-likelihood regions of both chosen and rejected responses, thereby biasing the model toward ambiguous outputs~\citep{ren2025learningdynamicsllmfinetuning,lv2026hidden}.

In this work, we propose \textbf{\mname}, a preference data filtering framework driven by intrinsic model signals. It integrates Mining Clearer Preference Samples via \underline{\textbf{Align}}ment Discrepancy and Sample \underline{\textbf{Diff}}iculty-aware Calibration to select a high-quality preference dataset. \mname separates two orthogonal concerns: \textit{preference-signal clarity}---whether a pair has a reliable preference direction---and \textit{sample difficulty}---how informative the pair is for the current model to learn from. Stage~1 uses Alignment Discrepancy ($R_{\mathrm{AD}}$) to address clarity by exploiting positive and inverse implicit reward margins; Stage~2 uses the Average Negative Log-Likelihood Gap (ANG) to address difficulty by prioritizing more informative pairs after polarity has been established.
Our key observation is that standard DPO can exploit positive and inverse preference signals by swapping chosen and rejected responses, thus providing complementary information. Motivated by this, we first filter raw preference pairs using $R_{\mathrm{AD}}$ to identify samples with clear preferences. 
However, some samples in this dataset are simplistic for the model, yielding negligible training benefit, and may even induce a “squeezing effect”~\citep{ren2025learningdynamicsllmfinetuning}: it further shrinks the high-likelihood regions of both chosen and rejected responses.
To address these issues, we calibrate the dataset through the lens of sample difficulty: estimating difficulty via the average negative log-likelihood gap and prioritizing more challenging samples. Extensive experiments demonstrate the effectiveness of \mname.
On AlpacaEval 2.0, \mname surpasses the strongest baseline SDPO~\citep{gao2025principleddataselectionalignment} by 6.3 and 3.2 points in length-controlled win rate on LLaMA-3-8B-SFT and Qwen2.5-7B-SFT, respectively, while maintaining comparable response lengths. The same trend holds on Arena-Hard and MT-Bench, where \mname achieves the strongest overall performance across both backbones. A difficulty-based curriculum built on the filtered data yields further gains.\footnote{Our code is publicly available at \url{https://github.com/sustech-nlp/AlignDiff}.}
\section{Background}
We first review the core ideas of Reinforcement Learning from Human Feedback (RLHF)~\citep{rlhf}, which naturally leads to the implicit reward mechanism employed in Direct Preference Optimization (DPO)~\citep{dpo}, the foundation of our method.

RLHF is a widely used framework for aligning LLMs with human preferences via preference learning. Let $\mathcal{D}$ be a dataset consisting of pairwise comparisons $(\mathbf{x}, \mathbf{y}_w, \mathbf{y}_l)$, where $\mathbf{y}_w$ is the preferred response over $\mathbf{y}_l$ for a given prompt $\mathbf{x}$, RLHF trains a reward model $r_{\phi}(\mathbf{y} | \mathbf{x})$ using the Bradley–Terry model~\citep{bt-model} and optimizes it by minimizing the cross-entropy loss. The policy $\pi_{\boldsymbol{\theta}}$ is refined via RL algorithms such as PPO~\citep{ppo} by maximizing the expected reward, with a KL constraint $\lambda$ enforcing the learned policy to stay close to the reference policy $\pi_{\text{ref}}$:
\begin{equation}
\vspace{-3.5pt}
\scalebox{0.9}{$
\begin{aligned}
\mathcal{J}_{\text{RLHF}}(\boldsymbol{\theta})
= {} & \mathbb{E}_{\mathbf{x} \sim \mathcal{D},\, \mathbf{y} \sim \pi_{\boldsymbol{\theta}}(\cdot | \mathbf{x})}
\left[ r_{\phi}(\mathbf{y} | \mathbf{x}) \right] \\
& {} - \lambda \cdot
\mathbb{D}_{\text{KL}}\!\left(
\pi_{\boldsymbol{\theta}}(\cdot | \mathbf{x})
\,\|\, 
\pi_{\text{ref}}(\cdot | \mathbf{x})
\right).
\end{aligned}$}
\vspace{-3.5pt}
\end{equation}
\begin{figure}
    \centering
    \includegraphics[width=1\linewidth]{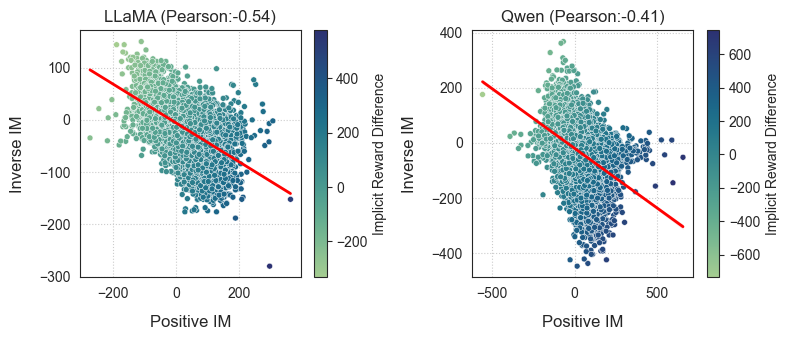}
    \vspace{-20pt}
    \caption{\textbf{{Complementarity between inverse and positive preference signals}.} We present the scatter plot of the joint distribution of positive and inverse implicit reward margins (IM) computed by SFT models on UltraFeedback samples. The results show that inverse IM and positive IM do not exhibit a clear inverse correlation, indicating that inverse signals can provide additional information not captured by positive signals, helping to understand the model’s preference distribution better.}
    \label{fig:im-vs-em}
\end{figure}
\begin{figure*}[ht]
    \centering
    \includegraphics[width=1\linewidth]{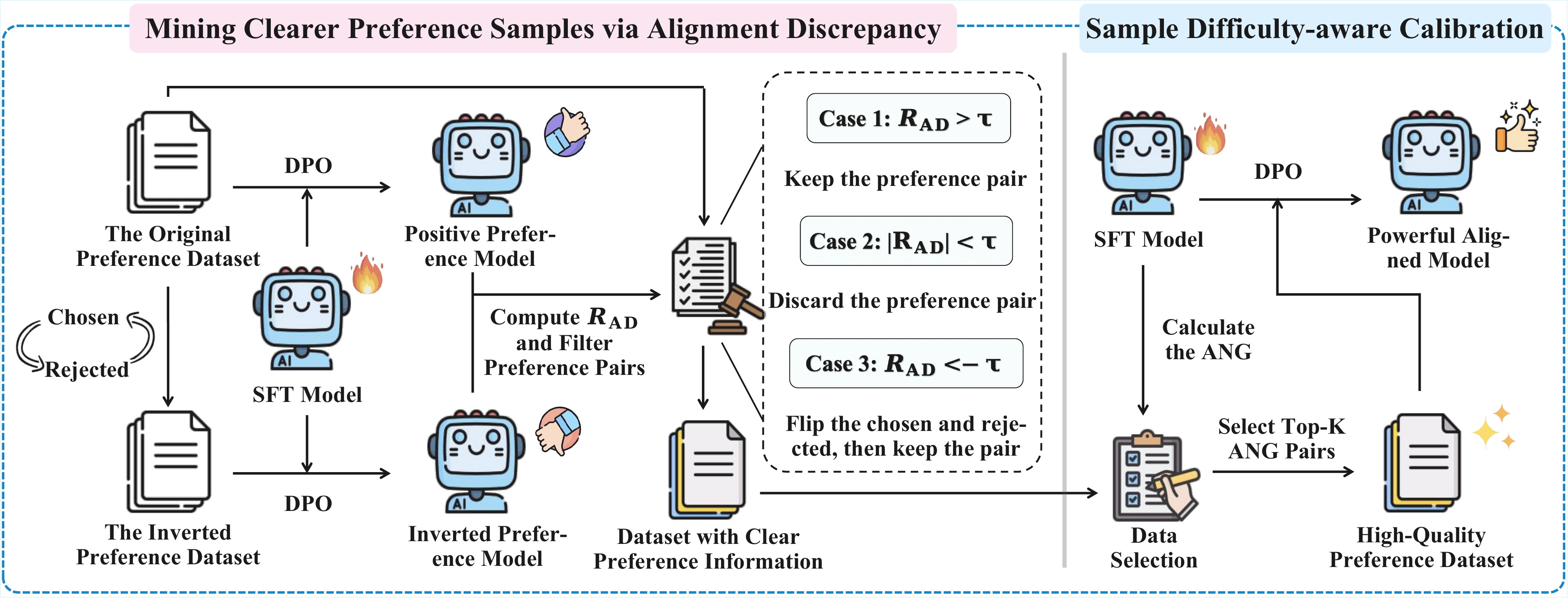}
    \vspace{-15pt}
    \caption{\textbf{{Overview of the \mname Framework}}. Stage~1 uses Alignment Discrepancy ($R_{\mathrm{AD}}$) with positive and inverse implicit reward margins to retain polarity-clear pairs; Stage~2 ranks them by the Average Negative Log-Likelihood Gap (ANG) to prioritize informative hard pairs and mitigate the squeezing effect. \mname relies solely on internal model signals.}
    \label{fig:framework-overall}
\end{figure*}
DPO simplifies the alignment process by removing the need for an explicit reward model~\citep{xue2026reason}. Instead, it defines an implicit reward using the policy model $\pi_{\boldsymbol{\theta}}$ and a reference model $\pi_{\text{ref}}$.
For a prompt $\mathbf{x}$ and response $\mathbf{y}$, the implicit reward is defined as:
\begin{equation}
\label{eqn:IM}
\resizebox{0.3\textwidth}{!}{ 
$r_{\text{im}}(\mathbf{y} | \mathbf{x};\pi_{\boldsymbol{\theta}}) = \log \frac{\pi_{\boldsymbol{\theta}}(\mathbf{y} | \mathbf{x})}{\pi_{\text{ref}}(\mathbf{y} | \mathbf{x})}.$}
\end{equation}
DPO models preferences between a chosen response $\mathbf{y}_w$ and a rejected response $\mathbf{y}_l$ based on the implicit reward margin:
\begin{equation}
\resizebox{0.4\textwidth}{!}{
$\mathrm{M}_{\mathrm{im}}(\mathbf{y}_w, \mathbf{y}_l | \mathbf{x}) = r_{\text{im}}(\mathbf{y}_w | \mathbf{x};\pi_{\boldsymbol{\theta}}) - r_{\text{im}}(\mathbf{y}_l | \mathbf{x};\pi_{\boldsymbol{\theta}}).$}
\end{equation}
The preference probability is then modeled as:
\begin{equation}
\label{eq:dpo-prefer-eq}
\resizebox{0.4\textwidth}{!}{
$p(\mathbf{y}_w \succ \mathbf{y}_l | \mathbf{x}) = \sigma \left( \beta \cdot \mathrm{M}_{\mathrm{im}}(\mathbf{y}_w, \mathbf{y}_l | \mathbf{x}) \right),$
}
\end{equation}
where $\beta$ controls the logit-scale sensitivity. To align the model, DPO optimizes a contrastive loss over the preference dataset $\mathcal{D}$, which corresponds to minimizing the negative log-likelihood of this preference probability:
\begin{equation}
\resizebox{0.4\textwidth}{!}{
$\mathcal{L}_{\text{DPO}} = - \mathbb{E}_{(\mathbf{x}, \mathbf{y}_w, \mathbf{y}_l) \sim \mathcal{D}} \Big[ \log p(\mathbf{y}_w \succ \mathbf{y}_l | \mathbf{x}) \Big].$}
\end{equation}
In essence, minimizing this loss is equivalent to maximizing the implicit reward margin $\mathrm{M}_{\mathrm{im}}$. A larger margin indicates clearer preference signals, which simplifies the optimization process, improves stability, and correlates with faster convergence \citep{dpo}. Consequently, several works \citep{dice,deng2025moreimprovingllmalignment} have proposed using IM as a core criterion to identify high-quality preference data.

\section{\mname}
We aim to extract a high-quality subset from a preference dataset. \mname (Figure~\ref{fig:framework-overall}) performs model-internal selection in two stages that target distinct questions: \textbf{Stage~1} ($R_{\mathrm{AD}}$) asks whether a pair has \textit{clear preference polarity}; \textbf{Stage~2} (ANG) asks, among polarity-clear pairs, which are \textit{most informative} for learning. Inspired by the symmetry of the DPO loss, we propose Alignment Discrepancy~(\S\ref{sec:ad}) to exploit positive and inverse implicit reward margins for polarity filtering. We then calibrate the dataset by sample difficulty~(\S\ref{sec:sample_difficulty}), retaining harder yet polarity-consistent pairs. The details of \mname are shown in Algorithm~1.
\subsection{Motivation}
We find that the IM is inherently tied to modeling positive preference signals in standard DPO. Interestingly, due to its symmetry, the DPO objective can also be used to model negative preference signals when the chosen and rejected responses are swapped in Eq. \ref{eq:dpo-prefer-eq}. This insight leads to a natural question: \textit{Can we leverage both positive and inverse preference signals to facilitate the selection of higher-quality preference data?} Formally, we define two policy models: (1) The positive preference policy $\pi_{\boldsymbol{\theta}}^{\text{pos}}$, trained to align with preferred responses via the standard DPO loss:
\begin{equation}
\label{eq:pos-dpo}
\resizebox{0.4\textwidth}{!}{$
\begin{aligned}
\mathcal{L}^{+}_{\mathrm{DPO}}&= - \mathbb{E}_{(\mathbf{x}, \mathbf{y}_w, \mathbf{y}_l) \sim \mathcal{D}} \Big[ 
\log \sigma \big(\\&
\beta \, r_{\text{im}}(\mathbf{y}_w | \mathbf{x};\pi_{\boldsymbol{\theta}}^{\text{pos}})
- \beta \, r_{\text{im}}(\mathbf{y}_l | \mathbf{x};\pi_{\boldsymbol{\theta}}^{\text{pos}})
\big)
\Big],
\end{aligned}$}
\vspace{-3pt}
\end{equation}
and (2) the inverse preference policy $\pi_{\boldsymbol{\theta}}^{\text{inv}}$, trained on inverse preference data to prefer typically dispreferred responses, with a DPO loss that reverses the preferred and dispreferred responses:
\begin{equation}
\label{eq:inv-dpo}
\resizebox{0.4\textwidth}{!}{$
\begin{aligned}
\mathcal{L}^{-}_{\mathrm{DPO}}&= - \mathbb{E}_{(\mathbf{x}, \mathbf{y}_w, \mathbf{y}_l) \sim \mathcal{D}} \Big[ 
\log \sigma \big(\\&
\beta \, r_{\text{im}}(\mathbf{y}_w | \mathbf{x};\pi_{\boldsymbol{\theta}}^{\text{inv}})
- \beta \, r_{\text{im}}(\mathbf{y}_l | \mathbf{x};\pi_{\boldsymbol{\theta}}^{\text{inv}})
\big)
\Big].
\end{aligned}$}
\end{equation}
These two policy models induce different reward behaviors: while $\pi_{\boldsymbol{\theta}}^{\text{pos}}$ encourages alignment with human preferences, $\pi_{\boldsymbol{\theta}}^{\text{inv}}$ exhibits systematically inverted tendencies. The differences between the implicit reward margins induced by the two models provide valuable signals for analyzing preference polarity and understanding model behavior. It can be observed from Figure~\ref{fig:im-vs-em} that the implicit reward margin $\mathrm{M}^{\mathrm{inv}}_{\mathrm{im}}(\mathbf{y}_w,\mathbf{y}_l|\mathbf{x})$ computed by $\pi_{\boldsymbol{\theta}}^{\text{inv}}$ does not exhibit a strong negative correlation with $\mathrm{M}^{\mathrm{pos}}_{\mathrm{im}}(\mathbf{y}_w,\mathbf{y}_l|\mathbf{x})$ computed by $\pi_{\boldsymbol{\theta}}^{\text{pos}}$. \textbf{This suggests that inverse preference signals are not merely the opposite of positive ones but imply that they capture complementary aspects of preference information.} As empirically validated in \S\ref{sec:ablation}, integrating both signals yields superior performance compared to relying on either signal in isolation.
\subsection{Mining Clearer Preference Samples via Alignment Discrepancy}
\label{sec:ad}
Motivated by the observation above, we define Alignment Discrepancy ($R_{\mathrm{AD}}$) as the difference between the margins $\mathrm{M}^{\mathrm{pos}}_{\mathrm{im}}$ and $\mathrm{M}^{\mathrm{inv}}_{\mathrm{im}}$, computed respectively from $\pi_{\boldsymbol{\theta}}^{\text{pos}}$ and $\pi_{\boldsymbol{\theta}}^{\text{inv}}$:
\begin{equation}
\label{eq:ad}
\scalebox{0.87}{$
R_{\mathrm{AD}}(\mathbf{y}_w,\mathbf{y}_l | \mathbf{x})
=\log\frac{\pi^{\mathrm{pos}}_{\boldsymbol{\theta}}(\mathbf{y}_w | \mathbf{x})}
          {\pi^{\mathrm{pos}}_{\boldsymbol{\theta}}(\mathbf{y}_l | \mathbf{x})}
-
\log\frac{\pi^{\mathrm{inv}}_{\boldsymbol{\theta}}(\mathbf{y}_w | \mathbf{x})}
          {\pi^{\mathrm{inv}}_{\boldsymbol{\theta}}(\mathbf{y}_l | \mathbf{x})}.$}
\end{equation}
The detailed derivation can be found in the Appendix~\ref{app:derivation-of-ad}. Compared to traditional methods relying only on positive signals, ${R_{\mathrm{AD}}}$ leverages implicit reward margins from both $\pi_{\boldsymbol{\theta}}^{\text{pos}}$ and $\pi_{\boldsymbol{\theta}}^{\text{inv}}$ to better capture preference consistency and detect anomalies. 
% It requires no extra annotations or complex features, is easy to compute, and effectively filters out mislabeled or abnormal samples, enhancing data quality and filtering efficiency. 
Therefore, based on the properties of $R_{\mathrm{AD}}$, we can perform a more refined filtering of the original data and fully leverage it. Specifically, we first define a labeling function $\phi$ that categorizes each sample into one of three preference types. Given a threshold $\tau > 0$, let $R \triangleq R_{\mathrm{AD}}(\mathbf{y}_w,\mathbf{y}_l|\mathbf{x})$. The preference label $\phi(R;\tau)$ is defined as:
\begin{equation}
\resizebox{0.33\textwidth}{!}{
$\phi(R;\tau)=
\begin{cases}
\operatorname{sign}(R), & |R| > \tau \\
0, & |R| \le \tau
\end{cases}\,\,,$}
\end{equation}
where the label $\phi$ encodes internal preference polarity based on $R_{\mathrm{AD}}(\mathbf{y}_w,\mathbf{y}_l|\mathbf{x})$:
$\phi = -1$ indicates a clear inverse preference, where the model favors the rejected response $\mathbf{y}_l$, potentially due to annotation noise or inconsistencies; $\phi = 0$ denotes an ambiguous case with weak or uncertain polarity; and $\phi = 1$ represents a clear positive preference, where the model’s choice aligns with the original human annotation. Given $\phi^{(i)}$, we can process each preference pair as follows:
samples with $\phi^{(i)}=0$ are discarded,
while those with $\phi^{(i)} \neq 0$ are retained with possibly reordered responses:
\begin{equation}
\resizebox{0.4\textwidth}{!}{$
(\mathbf{x}^{(i)}, \tilde{\mathbf{y}}_w^{(i)}, \tilde{\mathbf{y}}_l^{(i)}) =
\left\{
\begin{aligned}
(\mathbf{x}^{(i)}, \mathbf{y}_w^{(i)}, \mathbf{y}_l^{(i)}),\, & \phi^{(i)} = 1,\\
(\mathbf{x}^{(i)}, \mathbf{y}_l^{(i)}, \mathbf{y}_w^{(i)}),\, & \phi^{(i)} = -1.
\end{aligned}
\right.
$},
\end{equation}
Therefore, we can construct a higher-quality preference dataset $\tilde{\mathcal{D}} =
\Big\{
(\mathbf{x}^{(i)}, \tilde{\mathbf{y}}_w^{(i)}, \tilde{\mathbf{y}}_l^{(i)})
\;\Big|\;
\phi^{(i)} \neq 0
\Big\}.$

\begin{table*}[ht]
  \centering
  \caption{\textbf{{Performance comparison on AlpacaEval 2.0, Arena-Hard, and MT-Bench using DPO-trained models with various data subsets}}. The base models are LLaMA-3-8B-SFT and Qwen2.5-7B-SFT. The best-performing subsets are highlighted in \textbf{bold}, and the second-best ones are \underline{underlined}. }
  \vspace{-3pt}
  \resizebox{0.95\textwidth}{!}{
  \renewcommand{\arraystretch}{1.4}
  \fontsize{22pt}{20pt}\selectfont   % <-- 人为放大
  \begin{tabular}{l||ccccc|ccccc}
  \specialrule{1.5pt}{0pt}{0pt}
  \multirow{4}{*}{\textbf{Method}} 
  & \multicolumn{5}{c|}{\textbf{LLaMA-3-8B-SFT}} 
  & \multicolumn{5}{c}{\textbf{Qwen2.5-7B-SFT}} \\ 
  \cline{2-11}
  
  & \multicolumn{3}{c}{\textbf{Alpaca Eval 2.0}} 
  & \textbf{Arena-Hard} 
  & \textbf{MT-Bench} 
  & \multicolumn{3}{c}{\textbf{Alpaca Eval 2.0}} 
  & \textbf{Arena-Hard} 
  & \multicolumn{1}{c}{\textbf{MT-Bench}} \\ 
  \cline{2-11}
  
  & LC (\%) & WR (\%) & Len 
  & WR (\%) 
  & Score 
  & LC (\%) & WR (\%) & Len 
  & WR (\%) 
  & Score \\ 
  \hline
  
  Init 
  & 1.2 & 2.2 & 3,841 
  & 7.7 
  & 5.2$^{*}$ 
  & 5.3 & 4.7 & 1,230 
  & 10.7 
  & 5.7 \\
  
  Full 
  & 13.7 & 16.8 & 3,431 
  & 36.2 
  & 6.5$^{*}$ 
  & 21.3 & 18.9 & 1,715 
  & 46.9 
  & 6.8 \\
  
  PPLGAP 
  & 3.7 & 6.3 & 4,002 
  & 33.7 
  & 6.5 
  & 21.2 & 16.9 & 1,631 
  & 39.0
  & 6.7 \\
  
  LCPP 
  & 14.6 & 20.1 & 5,138 
  & 25.0
  & \underline{7.1} 
  & 20.0 & 22.6 & 2,050 
  & 38.8 
  & \underline{7.2} \\
  
  EM 
  & 12.6 & 13.5 & 2,076 
  & 33.5 
  & 7.0 
  & 21.7 & 19.4 & 1,740 
  & 46.6 
  & \underline{7.2} \\
  
  IM 
  & 19.1 & 19.8 & 2,428 
  & 38.9 
  & \underline{7.1} 
  & 27.8 & 27.0 & 1,928 
  & 57.0
  & \textbf{7.3} \\

  $\text{M}_{AP}$ 
  & 8.5 & 10.8 & 3375 
  & 17.3 
  & 6.4
  & 17.1 & 14.6 & 1,631 
  & 15.6
  & 6.9 \\
  
  IM\&EM 
  & 16.2 & 17.6 & 2,592 
  & 38.9 
  & \underline{7.1} 
  & 25.5 & 25.3 & 1,951 
  & 57.8 
  & \textbf{7.3} \\
  
  R.I.P 
  & 15.9 & 20.0 & 3,461 
  & 38.9 
  & \underline{7.1} 
  & 25.1 & 24.3 & 1,902 
  & \underline{55.8} 
  & \textbf{7.3} \\
  
  SDPO 
  & \underline{20.1} & \underline{21.3} & 2,501 
  & \underline{45.3} 
  & 6.9 
  & \underline{30.2} & \underline{28.8} & 1,902 
  & 55.3 
  & \underline{7.2} \\
  
  \specialrule{1.25pt}{0pt}{0pt}
  \textbf{\mname} 
  & \textbf{26.4} & \textbf{29.3} & 2,680 
  & \textbf{47.0} 
  & \textbf{7.2} 
  & \textbf{33.4} & \textbf{33.8} & 2,035 
  & \textbf{58.7} 
  & \textbf{7.3} \\
  
  \specialrule{1.5pt}{0pt}{0pt}
  \end{tabular}}
  \begin{flushleft}
  % \vspace{-2pt}
  \tiny
  \scriptsize Results marked with * in the table are from SDPO.
  \end{flushleft}
  \label{talbe:main_res}
  \vspace{-5pt}
  \end{table*}

\subsection{Sample Difficulty-Aware Dataset Calibration}
\label{sec:sample_difficulty}
Stage~1 ($R_{\mathrm{AD}}$) yields a preference dataset $\tilde{\mathcal{D}}$ with clear polarity, helping the model capture the fundamental alignment direction~\citep{dpo}. Stage~2 addresses a distinct question: \textit{which polarity-clear pairs are most informative to learn from?} Some retained samples remain too easy---the chosen response is already highly likely under the reference model while the rejected response is already unlikely---and thus offer limited training benefit~\citep{xue2026supervised}. Optimizing such pairs may further trigger the ``squeezing effect''~\citep{ren2025learningdynamicsllmfinetuning,pal2024smaugfixingfailuremodes,lan2025mappomaximumposterioripreference}: both chosen and rejected responses are pushed into overly narrow high-likelihood regions (Figure~\ref{fig:chosen_r} and~\ref{fig:rejected_r}), yielding ambiguous generations.

Importantly, we do \textit{not} treat difficulty as the absolute quality of a single response. DPO optimizes pairwise preference advantage, so we define difficulty as a \textit{relative contrast within a pair}: an informative pair should present a chosen response that remains challenging for the model to favor, alongside a rejected response that is comparatively easier to disfavor. This ``hard chosen, easy rejected'' structure is captured jointly by a single pairwise score rather than two independent absolute-quality filters.

To measure this contrast robustly, we propose the Average Negative Log-Likelihood Gap (ANG), which is less sensitive to response length than perplexity (PPL).
We define the average negative log-likelihood (AvgNLL) of a response $\mathbf{y}$ of length $T$ given input $\mathbf{x}$ as $\overline{\mathrm{NLL}}(\mathbf{y}) = -\tfrac{1}{T}\sum_{t=1}^{T}\log P(y_t|y_{<t}, \mathbf{x})$.
ANG is the gap between the AvgNLL of the chosen and rejected responses:
\begin{equation}
    \mathrm{ANG}(\mathbf{y}_w,\mathbf{y}_l) 
    =  \overline{\mathrm{NLL}}(\mathbf{y}_w)-\overline{\mathrm{NLL}}(\mathbf{y}_l) .
\end{equation}
A larger (positive) $\mathrm{ANG}$ means $\mathbf{y}_w$ is relatively harder to generate than $\mathbf{y}_l$ under the reference model---i.e., the pair is more informative for preference learning---not that $\mathbf{y}_w$ is intrinsically high-quality in isolation. Conversely, a smaller (or negative) $\mathrm{ANG}$ indicates an already easy pair with limited learning signal and higher squeezing risk. Because $R_{\mathrm{AD}}$-filtering already removes ambiguous pairs, ANG ranking is applied only to polarity-clear data. We select the top-$K$\% pairs with the largest $\mathrm{ANG}$ values to construct the final dataset $\mathcal{D}_{\text{final}}$. 

\section{Experiments}
\subsection{Experimental Settings}
\paragraph{Datasets and Base Models.}
Following previous works on preference optimization and LLM alignment~\citep{curry-dpo,zephyr}, we adopt a widely used preference dataset, namely UltraFeedback\_Binarized~\citep{ultrafeedback}.
{Instead of using off-the-shelf “Instruct” models, we initialize from SFT checkpoints trained on UltraChat~\citep{ding2023enhancing}, which helps reduce biases from heterogeneous instruction-tuning data across different Instruct releases and provides a consistent baseline for isolating the gains of our method.} Following prior work~\citep{simpo,gao2025principleddataselectionalignment}, we use LLaMA-3-8B-SFT\footnote{\url{https://huggingface.co/princeton-nlp/Llama-3-Base-8B-SFT}}~\citep{simpo}
 and Qwen-2.5-7B-SFT\footnote{\url{https://huggingface.co/glorgao/Qwen2.5-7B-SFT}}~\citep{gao2025principleddataselectionalignment}
 as reference models.

\paragraph{Baselines.} To comprehensively evaluate our method, we compare it against multiple strong baselines (See Appendix~\ref{app:baselines} for details), and all methods use their originally reported optimal settings:
\textbf{(1) External Reward Margin (EM)}~\citep{yasunaga2024almaalignmentminimalannotation,he2025airsystematicanalysisannotations}: We use Qwen2.5-72B-Instruct~\citep{qwen2.5} to score preference pairs across four dimensions, average the scores to obtain a final reward, and select samples with the largest reward margin.
\textbf{(2) Implicit Reward Margin (IM)}~\citep{dice,spa}: We use the SFT model as the reference model and train on the raw dataset. Then, we compute the IM for all data in this dataset using the SFT model, and select the data with the largest IM.
\textbf{(3) PPLGap}: Using PPL as a difficulty proxy, we select pairs with the largest chosen–rejected PPL gap.
\textbf{(4) Longest-Chosen Preference Pair (LCPP)}: Motivated by prior findings~\citep{shen2024rethinkingdataselectionsupervised}, we adopt pairs with the longest chosen responses as a baseline.
\textbf{(5) $\mathbf{\text{M}_{AP}}$}~\citep{huang2025larger}: This method integrates both margins to quantify the gap from the model's current implicit reward margin to the target explicit reward margin, thereby filtering the preference data.
\textbf{(6) External \& Implicit Reward Margin (IM\&EM)}: Following \citet{deng2025moreimprovingllmalignment}, we aggregate external and implicit reward margins and select samples with the largest combined margin.
\textbf{(7) R.I.P}~\citep{rip}: Selecting preference pairs with the longest rejected responses, while ensuring a sufficient external reward margin.
\textbf{(8) SDPO}~\citep{gao2025principleddataselectionalignment}:  SDPO identifies sample difficulty via six reference models, removes hard samples, and applies training only to easy ones.
\begin{figure*}[ht]
  \centering
  \begin{subfigure}[t]{0.32\textwidth}
    \centering
    \includegraphics[width=\textwidth]{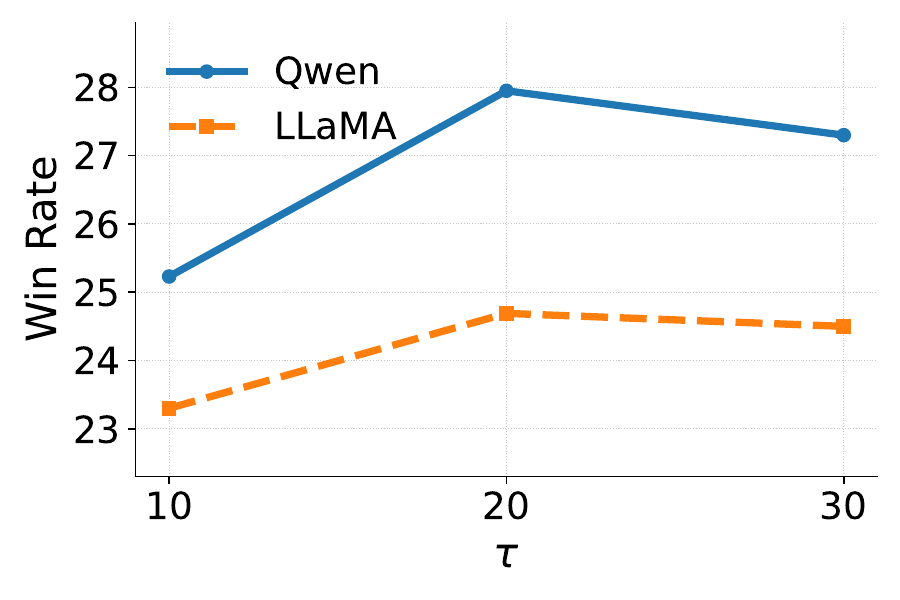}
    \vspace{-15pt}
    \caption{Hyperparameter $\text{$\tau$}$.}
    \label{fig:ablation_tau}
  \end{subfigure}
  \hfill
  \begin{subfigure}[t]{0.32\textwidth}
    \centering
    \includegraphics[width=\textwidth]{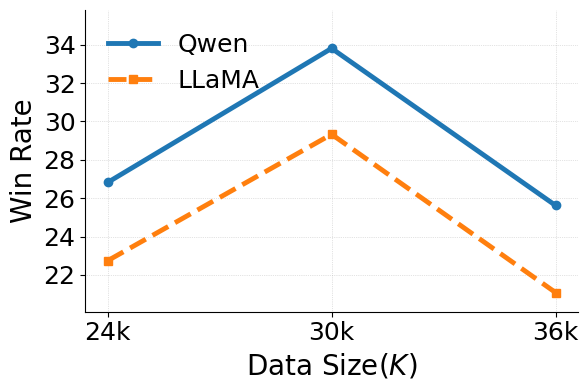}
    \vspace{-15pt}
    \caption{Hyperparameter $\text{K}$.}
    \label{fig:model_performance_vs_k}
  \end{subfigure}
  \begin{subfigure}[t]{0.32\textwidth}
    \centering
    \includegraphics[width=\textwidth]{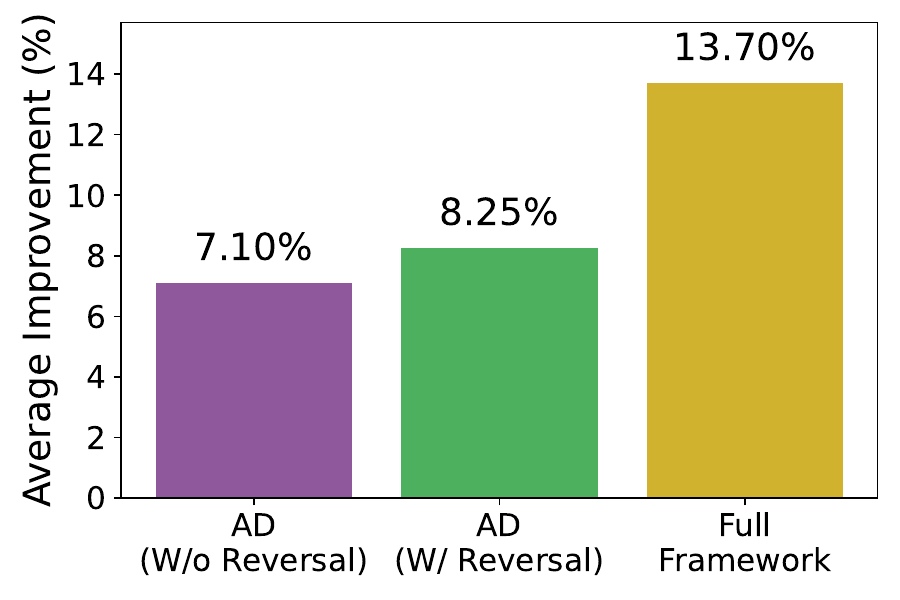}
    \vspace{-15pt}
    \caption{Avg. Improvement.}
    \label{fig:ablation_results}
  \end{subfigure}
  \vspace{-4pt}
  \caption{\textbf{{Results of Hyperparameter \& Ablation Studies}}. Panels (a) and (b) show the results of our hyperparameter studies; Panel (c) presents the ablation results for the role of each filtering step. }
  \vspace{-5pt}
\end{figure*}

\paragraph{Evaluation.} Following prior work~\citep{dice,beta-dpo,he2025airsystematicanalysisannotations}, we evaluate DPO-tuned models on three widely used benchmarks: 
(1) AlpacaEval 2.0~\citep{alpaca-eval} adopts a weighted win-rate evaluation protocol and reports both \textbf{WR} (win rate) and \textbf{LC} (length-controlled win rate) to assess response quality and alignment.
(2) Arena-Hard~\citep{arenahard2024} evaluates instruction-following ability and includes challenging real-world and creative tasks, with strong correlation to Chatbot Arena human preferences.
(3) MT-Bench~\citep{mtbench} assesses multi-turn dialogue capability to measure overall conversational performance. 
All evaluations follow the standard settings of each benchmark and use greedy decoding for generation.
For more detailed evaluation settings, please refer to the Appendix~\ref{app:evaluation}.

\paragraph{Implementation details.} 
All baselines use 30k samples for fair comparison, and all experiments are conducted under identical settings. More details of the training procedure are provided in Appendix~\ref{app:training_settings}. 
\subsection{Main Results}
\textbf{Implicit Rewards Consistently Outperform External Rewards.} Our main results are summarized in Table~\ref{talbe:main_res}. For both LLaMA-3-8B-SFT and Qwen2.5-7B-SFT, IM-filtered preference data consistently outperforms EM-filtered data on all three benchmarks. In contrast, IM\&EM, which incorporates external signals on top of IM, does not yield a clear win over IM alone. Relative to IM, IM\&EM reduces AlpacaEval LC and WR on both backbones (e.g., LLaMA-3-8B-SFT LC falls from 19.1\% to 16.2\%; Qwen2.5-7B-SFT LC from 27.8\% to 25.5\%), and on Arena-Hard ties IM on LLaMA-3-8B-SFT while showing only a small gain on Qwen2.5-7B-SFT (57.8\% vs.\ 57.0\%). A similar limitation is observed for $\mathrm{M}_{AP}$~\citep{huang2025larger}, which performs substantially worse than IM across both backbones. We conjecture that this is because $\mathrm{M}_{AP}$ only compares the absolute magnitudes of the target explicit and current implicit reward margins, while ignoring their directional consistency. As a result, incorrectly ordered pairs with large margins may still receive high scores, making the metric more sensitive to scale mismatch and anomalous or contradictory preference signals. These results suggest that effectively integrating external and internal information to achieve consistent performance gains remains challenging. This is consistent with evidence that LLM-as-a-judge scores can be driven by spurious correlations~\citep{zhang2026dibjudge,lai2026biasscopeautomateddetectionbias}; unlike some quality-aware settings that fuse an external estimator with an internal reward~\citep{wang2025marco}, adding EM on top of IM does not yield a clear gain for general-purpose alignment.

\textbf{\mname Outperforms Other Baselines.} \mname yields the strongest overall alignment among compared methods. With response lengths close to SDPO, LC exceeds SDPO by 6.3 points on LLaMA-3-8B-SFT and by 3.2 points on Qwen2.5-7B-SFT. The same pattern holds on Arena-Hard, where \mname improves win rates over SDPO by 1.7 points on LLaMA-3-8B-SFT and by 3.4 points on Qwen2.5-7B-SFT, and attains the best Arena-Hard scores among all filters on both backbones. Training LLaMA-3-8B-SFT on our top-50\% subset nearly doubles AlpacaEval~2.0 LC versus the full dataset (13.7\% $\rightarrow$ 26.4\%). We further assess the impact of our filtered data on WPO~\citep{zhou2024wpoenhancingrlhfweighted}, the strongest DPO variant from the SDPO paper. LLaMA-3-8B-SFT trained on the original dataset achieves a win rate of 19.37\%, whereas using our filtered data increases the win rate to 31.59\%, highlighting the substantial improvement brought by \mname. Taken together, these results underscore that \mname-filtered data can boost alignment performance, proving that higher data quality can indeed outweigh sheer quantity.
\begin{figure*}[ht]
  \centering
  \begin{subfigure}[t]{0.23\textwidth}
    \centering
    \includegraphics[width=\textwidth]{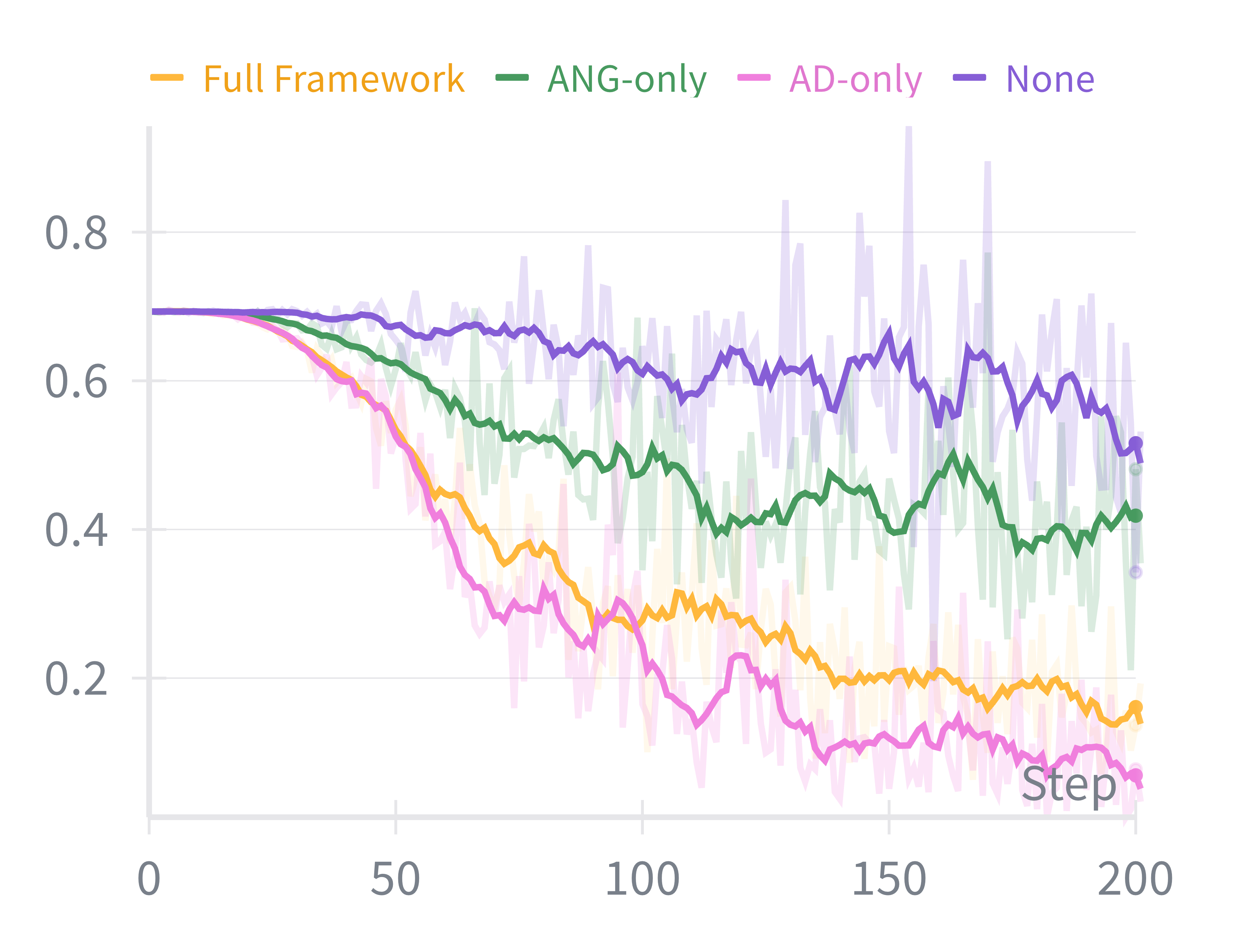}
    \vspace{-15pt}
    \caption{Training Loss}
    \label{fig:train_loss}
  \end{subfigure}
  \hfill
  \begin{subfigure}[t]{0.23\textwidth}
    \centering
    \includegraphics[width=\textwidth]{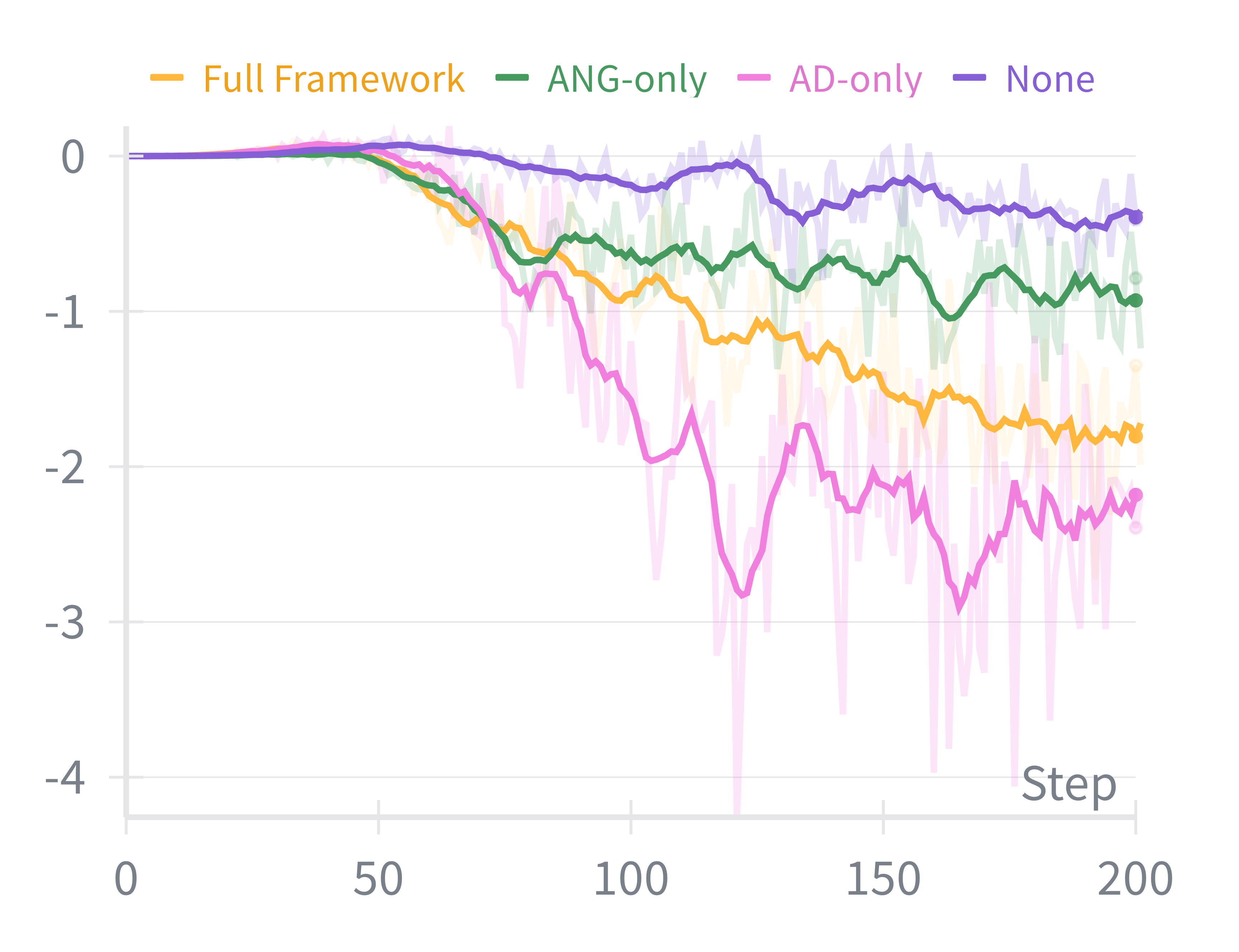}
    \vspace{-15pt}
    \caption{Chosen Rewards}
    \label{fig:chosen_r}
  \end{subfigure}
  \begin{subfigure}[t]{0.23\textwidth}
    \centering
    \includegraphics[width=\textwidth]{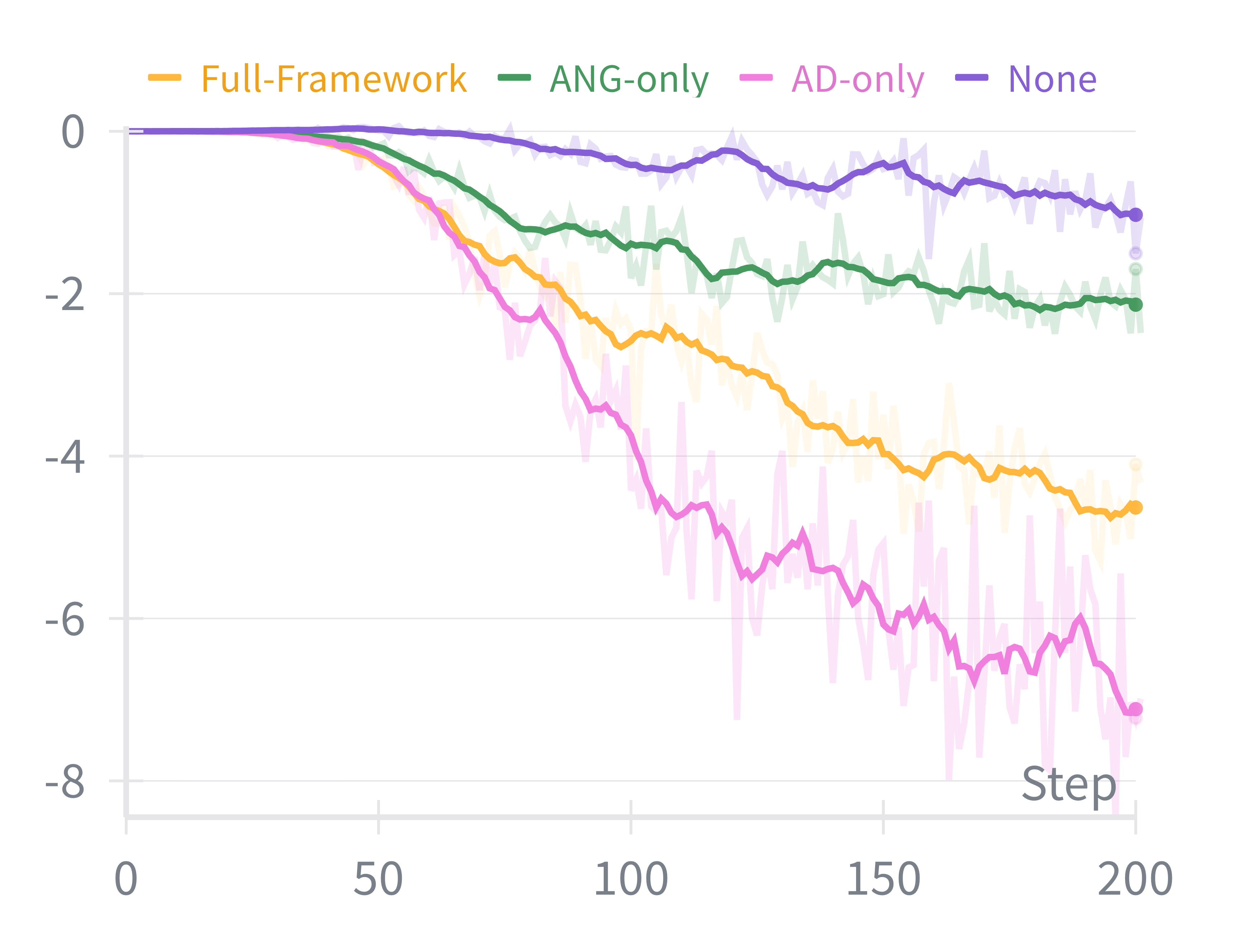}
    \vspace{-15pt}
    \caption{Rejected Rewards}
    \label{fig:rejected_r}
  \end{subfigure}
  \begin{subfigure}[t]{0.28\textwidth}
    \centering
    \includegraphics[width=\textwidth]{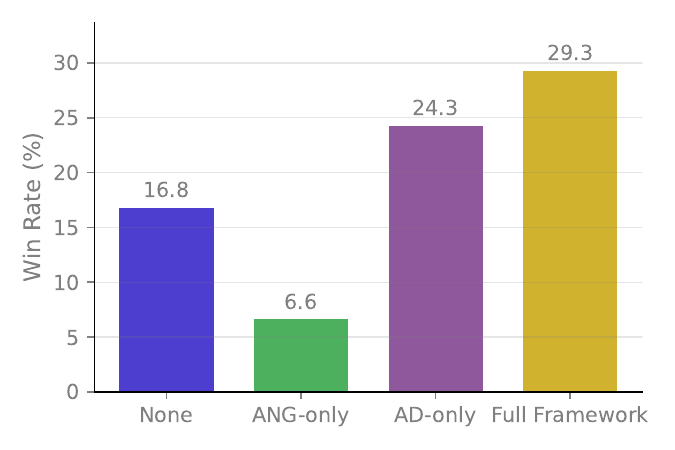}
    \vspace{-15pt}
    \caption{Performance Comparison}
    \label{fig:preformance_comparison}
  \end{subfigure}
  \vspace{-4pt}
  \caption{\textbf{{Training Dynamics and Performance}}. Dynamics of three metrics during DPO training: (a) training loss; (b) rewards of chosen responses; (c) rewards of rejected responses, each showing results for different filtering methods: None, $R_{\mathrm{AD}}$-only, ANG-only, and the full framework; and (d) performance comparison. Although $R_{\mathrm{AD}}$-only achieves lower loss, it is constrained by the squeezing effect; ANG-only mitigates squeezing but struggles to converge without polarity filtering.}
    \vspace{-5pt}
\end{figure*}
\subsection{Hyperparameter and Ablation Studies}
\label{sec:ablation}
\begin{figure}[ht]
    \centering
    \begin{subfigure}[t]{0.23\textwidth}
        \centering
        \includegraphics[width=\textwidth]{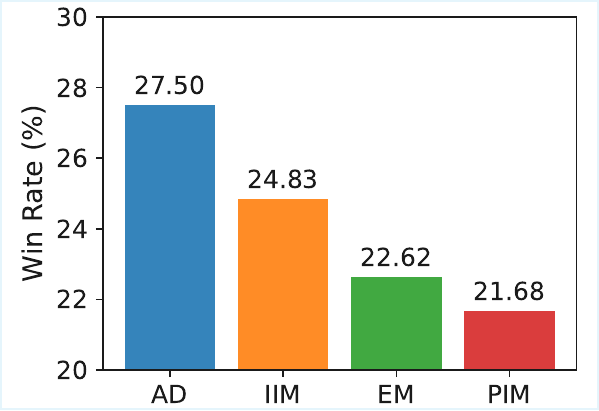}
        \caption{$R_{\mathrm{AD}}$ vs.\ other methods}
        \label{fig:ad_vs_others}
    \end{subfigure}
    \hfill
    \begin{subfigure}[t]{0.23\textwidth}
        \centering
        \includegraphics[width=\textwidth]{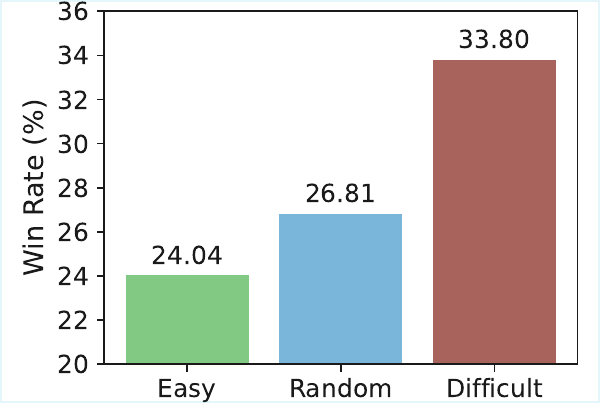}
        \caption{Calib. Strategy Comp.}
        \label{fig:stage2}
    \end{subfigure}
    \vspace{-4pt}
    \caption{\textbf{(a)} $R_{\mathrm{AD}}$-based filtering consistently outperforms PIM, IIM, and EM;
    \textbf{(b)} calibrating on hard samples boosts performance vs.\ easy/random baselines.}
    \label{fig:calibration}
    \vspace{-20pt}
\end{figure}

\textbf{Hyperparameters.} \mname introduces two key hyperparameters: \textbf{(1) $R_{\mathrm{AD}}$ threshold} $\tau>0$ (Figure~\ref{fig:ablation_tau}): for Qwen and LLaMA, setting $\tau=20$ achieves an optimal trade-off between data quality and quantity. Since the $R_{\mathrm{AD}}$ range may differ across models, we recommend selecting $\tau$ at approximately 5k-sample intervals and evaluating their effects.
\textbf{(2) Top-$K$ data selection} (Figure~\ref{fig:model_performance_vs_k}): choosing the top 30k samples ranked by ANG after $R_{\mathrm{AD}}$-filtering yields the best performance.
\textbf{Comparative Ablation of First-Stage Filtering Methods (Figure~\ref{fig:ad_vs_others})}. We employ $R_{\mathrm{AD}}$, positive IM (PIM), inverse IM (IIM), and EM for first-stage filtering, train DPO on the Qwen model with each resulting dataset, and compare performance. $R_{\mathrm{AD}}$ outperforms the other methods, demonstrating that combining positive and inverse IM provides richer signals for identifying pairs with clearer polarity.

\textbf{Validity of Calibration Using Difficult Samples (Figure~\ref{fig:stage2}).} We compare calibration strategies based on easy and random samples, and find that calibration based on difficult samples yields greater performance improvement than both. Among all settings, selecting easy samples for training results in the poorest performance, lagging behind the difficult-sample variant by 9.7\%. This gap suggests that challenging samples provide richer learning signals for model improvement.

\textbf{Role of Each Filtering Step (Figure~\ref{fig:ablation_results}).} To assess the contribution of each component in \mname, we conduct a step-by-step ablation study on the two models, incrementally adding key steps. We evaluate the following settings and compare the results with the original dataset: \textbf{(1) $R_{\mathrm{AD}}$ (W/o Reversal)}: applying $R_{\mathrm{AD}}$-based filtering to $\mathcal{D}$, retaining only samples with $\phi=1$; \textbf{(2) $R_{\mathrm{AD}}$ (W/ Reversal)}: applying both $R_{\mathrm{AD}}$-based filtering and the reversal operation, resulting in $\tilde{\mathcal{D}}$; and \textbf{(3) Full Framework}: applying the full pipeline, including ANG-based difficulty selection.
The results show that model performance consistently improves as we progressively incorporate components of \mname. Introducing $R_{\mathrm{AD}}$-based filtering without reversal yields a substantial improvement over the unfiltered baseline, and further adding the reversal operation brings additional gains. \mname achieves top performance by combining polarity clarity ($R_{\mathrm{AD}}$) and difficulty-aware selection (ANG), highlighting their complementarity.
\section{Analysis}
% Our ablation study assesses the impact of our method on model performance by incrementally adding key steps, though the underlying reasons remain unexplored.
% We conduct further analysis to provide a comprehensive understanding of the impact of the two stages in \mname. 
\subsection{Training Dynamics and Effectiveness}
\label{sec:training_dynamic}
\begin{figure}[!t]
\centering
\includegraphics[width=0.45\textwidth]{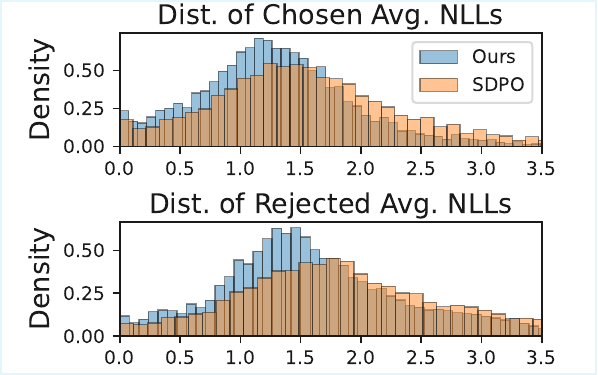}
    \vspace{-5pt}
  \caption{Dist. of Avg. NLLs. The model trained on filtered data exhibits a tighter, left-shifted distribution.}
  \label{fig:annls_dist}
  \vspace{-5pt}
\end{figure}

We examine training stability, reward quality, and model performance. We find that \textbf{(1) $R_{\mathrm{AD}}$ provides cleaner optimization signals} (Figure~\ref{fig:train_loss}): after incorporating $R_{\mathrm{AD}}$, the full framework trains more stably than ANG-only and converges to a level not far from $R_{\mathrm{AD}}$-only training. Although $R_{\mathrm{AD}}$-only achieves lower loss, the final performance (Figure~\ref{fig:preformance_comparison}) falls short of the full framework, illustrating the squeezing effect~\citep{ren2025learningdynamicsllmfinetuning,lv2026hidden}. \textbf{(2) Mitigating the squeezing effect requires difficulty-aware selection} (Figure~\ref{fig:chosen_r} and~\ref{fig:rejected_r}): learning polarity-clear pairs with relatively hard chosen and easy rejected responses stabilizes implicit rewards, prevents excessive deviation from the reference policy, and ensures relatively stable gradient signals. \textbf{(3) Combining $R_{\mathrm{AD}}$ and ANG enhances performance}~(Figure~\ref{fig:preformance_comparison}): without clear polarity, the model cannot learn effectively, yielding worse performance than no filtering; without informative hard pairs, performance is also limited by the squeezing effect. 
\textbf{(4) Leading to greater preference consistency (Figure~\ref{fig:annls_dist})}:
We compute the AvgNLLs of chosen and rejected samples on the small, high-quality preference dataset {argilla/dpo-mix-7k}\footnote{\url{https://huggingface.co/datasets/argilla/dpo-mix-7k}} via forced decoding, and compare the results with those of the best-performing baseline SDPO, which relies solely on implicit rewards. The results show that the model trained on our filtered data exhibits a left-shifted and sharper distribution for both chosen and rejected responses. This indicates that the model produces preferred responses more consistently. 

\begin{table}[ht]
\centering
\caption{Sample Ordering Strategies Comparison.}
\label{tab:curriculum_results}
\vspace{-5pt}
\resizebox{0.48\textwidth}{!}{
\begin{tabular}{lccc}
\toprule
\textbf{Sample Ordering Strategy} & \textbf{LC(\%)} & \textbf{WR(\%)} & \textbf{Length} \\
\hline
Random Order (baseline) &  33.38 & 33.81 &  2,035 \\
Hard-to-Easy            &  27.72 & 27.95&  2,005 \\
Easy-to-Hard            &  \textbf{36.40} &  \textbf{36.89} &  2,064 \\
\bottomrule
\end{tabular}}
\vspace{-15pt}
\end{table}
\subsection{Difficulty-based Curriculum Learning}
In this work, we adopt the ANG as a metric to measure sample difficulty, where larger gaps indicate harder examples. Building on the idea of curriculum learning used in DPO~\citep{curry-dpo}, we investigate whether a difficulty-based curriculum strategy can further improve model performance. We conduct an analytical experiment using the Qwen2.5-8B-SFT model, where the final 30k preference data selected by the whole \mname framework is organized according to three difficulty-based sample ordering strategies: \textbf{(1) random order}, serving as a baseline without any curriculum structure; \textbf{(2) hard-to-easy order}, where more difficult samples are presented to the model first; and \textbf{(3) easy-to-hard order}, where simpler samples are introduced first to enhance the model’s capability gradually. 
The results in Table~\ref{tab:curriculum_results} indicate significant performance differences under different training sample orders.
For the random order, LC/WR are 33.38/33.81; for the hard-to-easy order, they drop to 27.72/27.95; and for the easy-to-hard order, performance is the best, with LC/WR increasing to 36.40/36.89. Therefore, the easy-to-hard training strategy can further improve model performance, and the training order has an important impact on the outcome.
\subsection{Computational Efficiency Comparison}
To evaluate the trade-off between computational cost incurred during data selection and model performance, we conduct experiments and compare \mname with several baselines, including IM, EM, and SDPO. We report the GPU hours required for each method and compute efficiency as performance (average WR) per GPU hour. Please note that the computation of EM relies on the Qwen2.5-72B model for scoring across four dimensions, which leads to a relatively high computational cost. The detailed calculation of GPU hours for these methods can be found in the Appendix~\ref{app:cal_gpu_hours}. Table~\ref{tab:cost_tradeoff} shows that, compared with EM and SDPO, our method achieves a better balance between efficiency and performance. Although its efficiency is lower than IM's, our method attains superior performance while mitigating the squeeze effect. Moreover, we also explored a strategy where the first stage uses only half of the original dataset while keeping the second stage unchanged (referred to as Random-Half). The results indicate that although this strategy underperforms the full-data version, it surpasses the strongest baseline SDPO and attains efficiency comparable to IM (the most efficient method).
\begin{table}
% \vspace{-15pt}
\centering
\caption{Efficiency Comparison across Methods.}
\label{tab:cost_tradeoff}
\vspace{-5pt}
\resizebox{0.48\textwidth}{!}{
\begin{tabular}{lccc}
\toprule
\textbf{Method} & \textbf{GPU Hours} & \textbf{Avg. WR} & \textbf{Efficiency} \\ \hline
IM & 43 & 24.6 & 0.57 \\
EM & 128 & 17.2 & 0.13 \\
SDPO & 162 & 25.6 & 0.16 \\
\mname & 80.5 & 30.2 & 0.38 \\
\mname (Random-Half) & 48.5 & 28.1 & \textbf{0.58} \\	
\bottomrule
\end{tabular}}
\end{table}

\section{A Close Look at the Filtered High-Quality Data}
To better understand the nature of the high-quality preference data selected by \mname, we conduct a detailed analysis of the resulting dataset across three different models (LLaMA, Mistral, and Qwen).
\begin{figure*}[ht]
  \centering
  \begin{subfigure}[t]{0.32\textwidth}
    \centering
    \includegraphics[width=\textwidth]{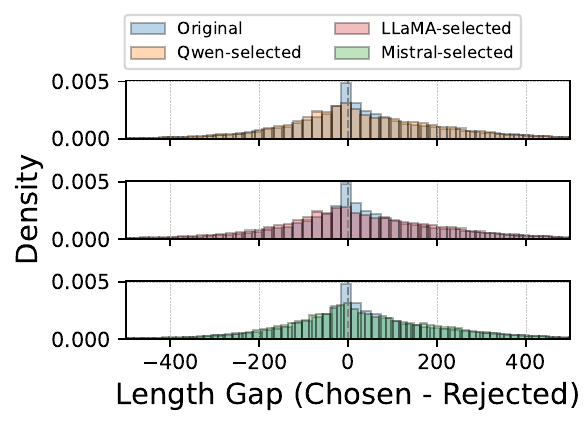}
    \vspace{-15pt}
    \caption{Length gap distribution}
    \label{fig:len_gap}
  \end{subfigure}
  \hfill
  \begin{subfigure}[t]{0.32\textwidth}
    \centering
    \includegraphics[width=\textwidth]{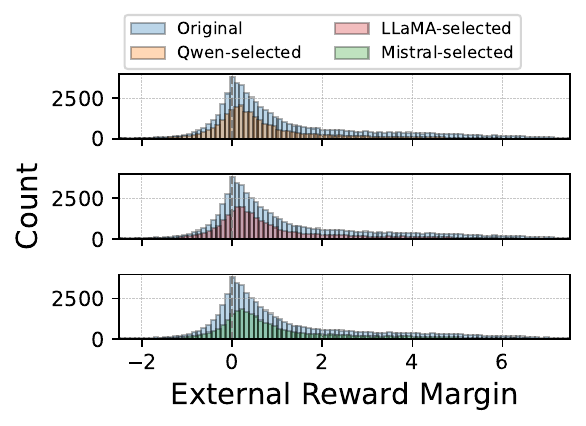}
    \vspace{-15pt}
    \caption{EM distribution}
    \label{fig:em_distribution}
  \end{subfigure}
  \begin{subfigure}[t]{0.32\textwidth}
    \centering
    \includegraphics[width=\textwidth]{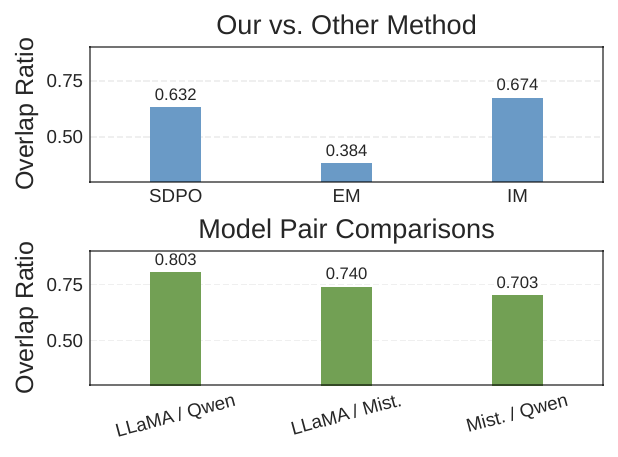}
    \vspace{-15pt}
    \caption{Overlap analysis}
    \label{fig:overlap_ratio}
  \end{subfigure}
  \vspace{-4pt}
  \caption{\textbf{{External Reward margin optimization with length gap preserved ((a) and (b)).}} The filtered data caused almost no change in the length gap distribution, while optimizing the external reward margin distribution, reducing ambiguous samples, and shifting the overall distribution slightly to the right. Panel (c) shows the comparative analysis from method and model Perspectives.}
  \label{fig:em_and_len_gap}
\end{figure*}

\textbf{Optimizing external reward margins while preserving length gap distribution.} In the filtered high-quality preference data, the distribution of length gaps closely matches that of the original dataset (Figure~\ref{fig:len_gap}), indicating that the improvement in data quality is not due to increased response length. Although our filtering relies solely on internal information, observing the external reward margin distribution (Figure~\ref{fig:em_distribution}) before and after filtering reveals that, for external rewards, the primarily removed samples are still ambiguous ones, and the distribution shifts slightly to the right.

\textbf{Comparative Overlap Analysis: Model-Level and Method-Level Perspectives.} At the method level (Figure~\ref{fig:overlap_ratio} (bottom)), our approach shows high overlap with IM (67.4\%) and SDPO (63.2\%) due to their shared reliance on internal model signals for selecting high-reward-margin samples. Nevertheless, \mname can still identify more representative and higher-quality data beyond the overlapping subset. In contrast, the overlap with EM is significantly lower (38.4\%) because EM relies on external models for multi-dimensional fine-grained scoring, making its selection criteria fundamentally different from internal-signal-based methods. This highlights the distinctions and complementarity between the two types of methods.
At the model level (Figure~\ref{fig:overlap_ratio} (top)), the high overlap rates indicate consistency among different model architectures in identifying high-quality samples, possibly stemming from their shared capabilities in language understanding and data quality assessment, which also validates the robustness of our method across different architectures. To further understand the properties of our scoring metric, we also show the distribution of $R_{\mathrm{AD}}$ for three SFT models on the UltraFeedback dataset in Figure~\ref{fig:rad_distribution}.

\section{Conclusion}
We propose \mname, a two-stage framework for filtering preference data using only the model’s internal signals. 
Our approach effectively removes ambiguous or conflicting samples by leveraging both positive and inverse preference signals through bidirectional alignment discrepancies. 
We further refine data selection using the average negative log-likelihood gap to identify high-quality, informative pairs. Both stages depend entirely on the model’s internal evaluation signals, thereby avoiding potential distributional shifts introduced by external reward functions and yielding data that better matches the model’s intrinsic learning tendencies. 
Experiments across multiple base LLMs and benchmarks show that \mname consistently outperforms strong preference data selection baselines while maintaining competitive response lengths, and that the filtered data further benefits from a difficulty-based curriculum. Overall, our results suggest that intrinsic model signals provide an effective foundation for data-centric alignment. In future work, it would be interesting to extend this framework to larger-scale models and broader post-training settings.

\section*{Acknowledgements}

This project was supported by the National Key R\&D Program of China (No. 2025YFB4007600), National Natural Science Foundation of China (No. 62306132), Guangdong Basic and Applied Basic Research Foundation (No. 2025A1515011564), Natural Science Foundation of Shanghai (No. 25ZR1402136). We thank the anonymous reviewers for their insightful feedback on this work. We used large language models only for language polishing and grammatical correction. All scientific content, analyses, and conclusions are solely our own, and we assume full responsibility for the integrity of this work.

\section*{Limitation}

\mname relies on intrinsic model signals (e.g., likelihood), which may vary with model scale and architecture, so it may be necessary to adjust the corresponding hyperparameters for different models. Moreover, although \mname demonstrates promising improvements, the filtering process inevitably discards part of the training data, which may also exclude some valuable but subtle preference signals. Our experiments are conducted on a single preference dataset (UltraFeedback\_Binarized) and limited to 7--8B-scale base models due to computational constraints, leaving the scalability of \mname to larger models and additional preference corpora as an open question. These aspects constitute important directions for future work.

\bibliography{reference}

\appendix

\section{Societal Impacts}
Aligning LLMs with human preferences is crucial for ensuring AI safety. By effectively filtering out inherent noise and toxic behaviors using intrinsic model signals, our framework, \mname, helps downstream algorithms better capture benign human preferences, thereby mitigating the risk of models generating harmful or biased content. Furthermore, \mname demonstrates that achieving superior alignment with a smaller, filtered dataset drastically reduces the carbon footprint and compute barriers, democratizing high-performance alignment for resource-constrained researchers. While relying on base model signals introduces a potential risk of amplifying data selection biases, this can be mitigated by auditing the cultural diversity of the filtered subsets and integrating external safety guidelines.

\section{Related Work}
\subsection{Preference Learning for Alignment}
Preference learning has been surveyed from a preference-centered view covering feedback sources, modeling, and usage~\citep{jiang2024survey,xiao2026bridgingrewardgenerationgapdirect}. Early work~\citep{instructiontuninggpt4,pkusaferlhf,qwen1} predominantly employs the RLHF~\citep{rlhf} framework, integrating algorithms such as PPO~\citep{ppo} with explicit reward modeling to improve alignment between models and human preferences. DPO~\citep{dpo} has been proposed as an efficient alternative approach that leverages implicit reward signals to directly optimize models using preference data, thereby streamlining the training process and substantially reducing computational costs. Recent analyses further interpret DPO as contrastive mutual-information maximization and attribute the late-stage decline of chosen likelihood to this formulation~\citep{lv2026hidden}. Building on DPO, more advanced preference learning algorithms such as IPO~\citep{ipo}, KTO~\citep{kto}, and SimPO~\citep{simpo} have emerged, aiming to improve the efficiency of preference modeling and enhance the alignment performance of language models. Recent studies further show that SFT can suffer from incomplete learning~\citep{xue2026supervised}, and that model-internal uncertainty provides an efficient signal for reward modeling~\citep{xue2026reason}. However, these algorithms all rely on high-quality preference data, and ensuring the quality of such data has become one of the key bottlenecks in further improving alignment performance.

\subsection{Data Selection in Preference Optimization}
% Selecting high-quality preference data poses several challenges, including annotation noise~\citep{r-dpo}, imbalanced preference distributions~\citep{chidambaram2024directpreferenceoptimizationunobserved}, and potential systemic biases~\citep{survey-dpo}. 
The alignment performance of a large language model is largely dependent on the quality of the preference data. Reliable preference optimization also depends on accurately capturing subtle semantic distinctions between candidate responses~\citep{xue2023dual,xue2025structcoh}. Early preference optimization work~\citep{llama2,ultrafeedback,helpsteer,llama3} adopts data selection techniques inherited from the pre-training and instruction-tuning stages, such as deduplication, quality classifiers, and heuristic filtering, while quality-aware methods further exploit mixed-quality instruction data rather than treating all samples equally~\citep{wang2024openchat}. These approaches typically rely on an external reward signal, which refers to the scores assigned by a reward model or an LLM-as-a-judge to evaluate candidate responses, and apply rejection sampling to construct preference pairs. Diversified annotator preferences can further degrade reward calibration~\citep{zeng2024diversified}, and LLM-as-a-judge scores may rely on spurious cues rather than genuine preference~\citep{zhang2026dibjudge}. Recent studies~\citep{beta-dpo,rs-dpo} suggest that, compared to methods that evaluate and filter individual responses based on absolute quality, selecting response pairs according to their reward gap is more beneficial for preference optimization, as it provides stronger training signals for alignment. Inspired by this, \citet{f-dpo,hu2024comprehensivepreferencedatacollection} leveraged the external reward gap to filter preference data. Unlike using external reward signals, \citet{kim2025spreadpreferenceannotationdirect} trained a weakly aligned model to compute implicit rewards and use the implicit reward gap (derived from the logit differences in the DPO objective and reflecting the model’s internal preferences) as the basis for data selection. \citet{qiyuan2025efficient} similarly re-rank preference pairs with the model’s own judgments to mitigate off-policy distribution shift. \citet{deng2025moreimprovingllmalignment} found that the external reward gap and the implicit reward gap exhibit a notably weak correlation; thus, they combined both the external reward gap and the implicit reward gap to refine the preference pair selection process. Combining external quality estimates with internal rewards has also been explored in quality-aware preference optimization~\citep{wang2025marco}. A parallel line selects training data using model-internal features rather than external rewards~\citep{shi2026igds}. Overall, this phenomenon reflects a paradigm shift in preference data selection, gradually transitioning from reliance on external reward signals to reliance on internal ones.
% \section{Hyperparameter study on the alignment discrepancy threshold of the Mistral model}
% \begin{figure}
%         \centering
%         \vspace{-10pt}
%         \includegraphics[width=1\linewidth]{figures/model_performance_vs_tau_mistral.pdf}
%         \caption{Hyperparameter Study $\tau$ for the Mistral Model.}
%         \label{fig:tau_mistral}
% \end{figure}
% In this section, we present a hyperparameter study on the alignment discrepancy (AD) threshold $\tau$ for the Mistral model. Since the AD values computed by different models vary, it is necessary to analyze the threshold separately for each model. To balance the threshold $\tau$ and the corresponding data volume, we select three $\tau$ values: 60, 80, and 100 for the Mistral model. The reason for this choice is that when $\tau$ is set too small (e.g., $\tau$ = 20), the amount of data filtered out is too limited, resulting in negligible differences in training performance. We aim for each $\tau$ to correspond to about 5k data points to ensure the effectiveness of the analysis. As shown in Figure~\ref{fig:tau_mistral}, setting $\tau=80$ achieves an optimal trade-off
% between data quality and quantity for Mistal.
\begin{table*}[ht]
  \centering
  \caption{\textbf{Average performance comparison of DPO-trained models on AlpacaEval 2.0 and Arena-Hard using different data subsets, evaluated with multiple judge models.}. The base models are LLaMA-3-8B-SFT and Qwen2.5-7B-SFT. The best-performing results are highlighted in \textbf{bold}, and the second-best ones are \underline{underlined}.}
  \vspace{-3pt}
  \resizebox{0.9\textwidth}{!}{
  \renewcommand{\arraystretch}{1.4}
  \fontsize{22pt}{20pt}\selectfont
  \begin{tabular}{l||cccc|cccc}
  \specialrule{1.5pt}{0pt}{0pt}
  \multirow{3}{*}{\textbf{Method}} 
  & \multicolumn{4}{c|}{\textbf{LLaMA-3-8B-SFT}} 
  & \multicolumn{4}{c}{\textbf{Qwen2.5-7B-SFT}} \\ 
  \cline{2-9}
  & \multicolumn{3}{c}{\textbf{AlpacaEval 2.0}} 
  & \textbf{Arena-Hard} 
  & \multicolumn{3}{c}{\textbf{AlpacaEval 2.0}} 
  & \textbf{Arena-Hard} \\ 
  \cline{2-9}
  & LC (\%) & WR (\%) & Len 
  & WR (\%) 
  & LC (\%) & WR (\%) & Len 
  & WR (\%) \\ 
  \hline

  Init 
  & 5.6  & 10.4 & 3,841 
  & 12.3 
  & 8.7  & 7.5  & 1,230 
  & 13.0 \\

  Full 
  & 20.9 & 28.0 & 3,431 
  & 31.5 
  & 31.1 & 26.9 & 1,715 
  & 40.0 \\

  PPLGAP 
  & 9.8  & 20.7 & 4,002 
  & 26.6 
  & 27.3 & 22.2 & 1,631 
  & 33.0 \\

  LCPP 
  & 22.1 & \underline{33.8} & 5,138 
  & 27.9 
  & 29.5 & 30.8 & 2,050 
  & 34.0 \\

  EM 
  & 20.0 & 25.1 & 2,076 
  & 30.9 
  & 31.8 & 27.7 & 1,740 
  & 36.3 \\

  IM 
  & 25.4 & 27.4 & 2,428 
  & 33.3 
  & 35.0 & 33.7 & 1,928 
  & \underline{42.5} \\

  IM\&EM 
  & 24.5 & 28.2 & 2,592 
  & 31.8 
  & 33.4 & 32.8 & 1,951 
  & 45.1 \\

  R.I.P 
  & 22.3 & 27.6 & 3,461 
  & 32.0 
  & 32.2 & 30.8 & 1,902 
  & 43.5 \\

  SDPO 
  & \underline{28.6} & 31.3 & 2,501 
  & \underline{34.6} 
  & \underline{36.7} & \underline{34.9} & 1,902 
  & 43.4 \\

  \specialrule{2pt}{0pt}{0pt}
  \textbf{AlignDiff} 
  & \textbf{29.5} & \textbf{34.9} & 2,680 
  & \textbf{36.4} 
  & \textbf{38.8} & \textbf{39.1} & 2,035 
  & \textbf{45.3} \\

  \specialrule{1.5pt}{0pt}{0pt}
  \end{tabular}}
  \label{table:additional_res}
\end{table*}
\begin{table*}[ht]
\centering
\caption{\textbf{{ Performance Comparison of downstream tasks (GPQA / Toxigen / TruthfulQA / MMLU / Winogrande) Using DPO-Trained Models with Various Data Subsets}}.}
\vspace{-3pt}
\label{tab:add_res}
\resizebox{0.8\textwidth}{!}{
\begin{tabular}{clcccccc}
\hline
Method &  & GPQA & Toxigen & TruthfulQA & MMLU & Winogrande & AVG \\ \hline
\multicolumn{8}{c}{LLaMA-3-8B-SFT} \\ \hline
PPLGAP &  & 0.328 & 0.427 & 0.493 & 0.613 & 0.729 & 0.518 \\
EM &  & 0.313 & 0.430 & 0.521 & 0.624 & 0.719 & 0.521 \\
IM &  & 0.317 & 0.429 & 0.488 & 0.622 & 0.719 & 0.515 \\
IM \& Em &  & 0.308 & 0.428 & 0.490 & 0.623 & 0.717 & 0.513 \\
LCPP &  & 0.332 & 0.427 & 0.498 & 0.612 & 0.730 & 0.520 \\
R.I.P &  & 0.324 & 0.427 & 0.492 & 0.620 & 0.721 & 0.517 \\
SDPO &  & 0.319 & 0.430 & 0.485 & 0.620 & 0.717 & 0.514 \\
\mname &  & 0.335 & 0.428 & 0.493 & 0.621 & 0.722 & 0.520 \\ \hline
% \multicolumn{8}{c}{Mistral-7B-SFT} \\ \hline
% PPLGAP &  & 0.319 & 0.559 & 0.528 & 0.571 & 0.750 & 0.545 \\
% EM &  & 0.304 & 0.587 & 0.619 & 0.584 & 0.738 & 0.566 \\
% IM &  & 0.304 & 0.579 & 0.620 & 0.571 & 0.729 & 0.560 \\
% IM \& Em &  & 0.306 & 0.572 & 0.635 & 0.566 & 0.741 & 0.564 \\
% LCPP &  & 0.308 & 0.476 & 0.496 & 0.583 & 0.740 & 0.520 \\
% R.I.P &  & 0.315 & 0.577 & 0.577 & 0.582 & 0.739 & 0.558 \\
% SDPO &  & 0.301 & 0.578 & 0.562 & 0.578 & 0.732 & 0.550 \\
% \mname &  & 0.317 & 0.572 & 0.586 & 0.582 & 0.740 & 0.559 \\ \hline
\multicolumn{8}{c}{Qwen2.5-7B-SFT} \\ \hline
PPLGAP &  & 0.339 & 0.567 & 0.611 & 0.710 & 0.720 & 0.589 \\
EM &  & 0.346 & 0.569 & 0.615 & 0.709 & 0.710 & 0.590 \\
IM &  & 0.344 & 0.569 & 0.603 & 0.709 & 0.707 & 0.586 \\
IM \& Em &  & 0.344 & 0.569 & 0.603 & 0.710 & 0.706 & 0.586 \\
LCPP &  & 0.326 & 0.569 & 0.602 & 0.709 & 0.707 & 0.583 \\
R.I.P &  & 0.348 & 0.568 & 0.608 & 0.709 & 0.707 & 0.588 \\
SDPO &  & 0.344 & 0.569 & 0.603 & 0.709 & 0.710 & 0.587 \\
\mname &  & 0.342 & 0.569 & 0.609 & 0.709 & 0.708 & 0.587 \\ \hline
\end{tabular}}
\end{table*}
\section{Additional Experimental Results}
\textbf{Evaluation results from multiple evaluators.} In our main experiments, we primarily rely on a single judge model for evaluation. This may raise a concern: whether the observed performance gains are influenced by biases of a specific judge, thereby calling into question the reliability of the results. To address this issue, we further conducted multi-judge evaluations on Arena-Hard (using GPT-4o-mini, GPT-4.1-nano, and Qwen3-Max as judgers) as well as on Alpaca-Eval 2 (using GPT-4o-mini, GPT-4.1-nano, and DeepSeek-V3 as judgers) to verify the robustness of our results. We reported the averaged performance across them in Table~\ref{table:additional_res}. As shown in Table~\ref{table:additional_res}, our method \textbf{\mname} consistently achieves the best performance across both base models and all evaluation metrics under multi-judge evaluation. Specifically, on LLaMA-3-8B-SFT, /mname attains the highest LC (29.5\%), WR (34.9\%), and Arena-Hard WR (36.4\%) on AlpacaEval 2.0 and Arena-Hard, respectively, outperforming the strongest baseline SDPO by 0.9\%, 3.6\%, and 1.8\% absolute points. On Qwen2.5-7B-SFT, /mname similarly leads with LC of 38.8\%, WR of 39.1\%, and Arena-Hard WR of 45.3\%, surpassing SDPO by 2.1\%, 4.2\%, and 1.9\% absolute points. These consistent gains across multiple judge models (GPT-4o-mini, GPT-4.1-nano, Qwen3-Max for Arena-Hard; GPT-4o-mini, GPT-4.1-nano, DeepSeek-V3 for AlpacaEval 2.0) demonstrate that the superiority of /m na me is not an artifact of any single judge's bias, but reflects genuine and robust improvements in alignment quality.

\textbf{Downstream task results.} We further evaluate the performance of DPO models trained with different baselines on diverse benchmarks (GPQA / Toxigen / TruthfulQA / MMLU / Winogrande) to assess their capabilities on downstream tasks, including question answering~\citep{xue2024question} (Table~\ref{tab:add_res}). Interestingly, under comparable data scales, different filtering strategies yield similar performance on downstream tasks. This phenomenon is consistent with the alignment tax effect~\citep{askell2021generallanguageassistantlaboratory}, which suggests that improvements in alignment or preference optimization do not necessarily translate into proportional gains on downstream performance. From this perspective, the limited performance differences across filtering strategies may indicate that downstream performance is influenced not only by data selection quality but also by inherent trade-offs introduced during the alignment training process. Nevertheless, our method consistently achieves competitive results, suggesting that even under a potential alignment tax, improving the effectiveness of data selection remains important.

\section{Pseudocode of \mname}
\begin{figure*}[ht]
    \centering
    \includegraphics[width=1\linewidth]{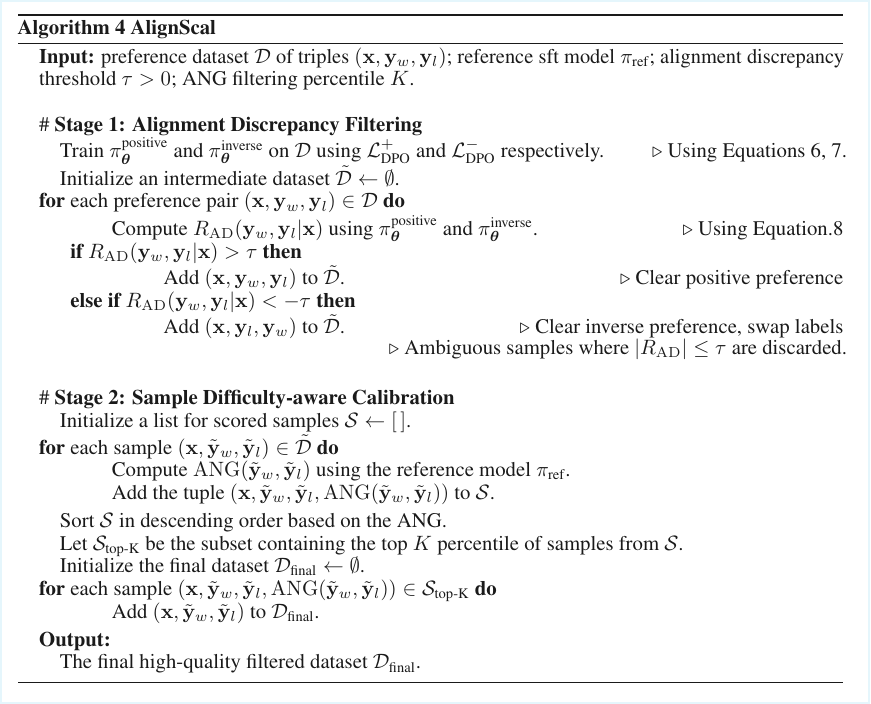}
    \vspace{-20pt}
    \caption{Pseudocode of \mname.}
    \label{fig:app_algo}
\end{figure*}

We show the detailed pseudocode of \mname in Figure~\ref{fig:app_algo}.

\begin{figure*}
    \centering
    \includegraphics[width=1\linewidth]{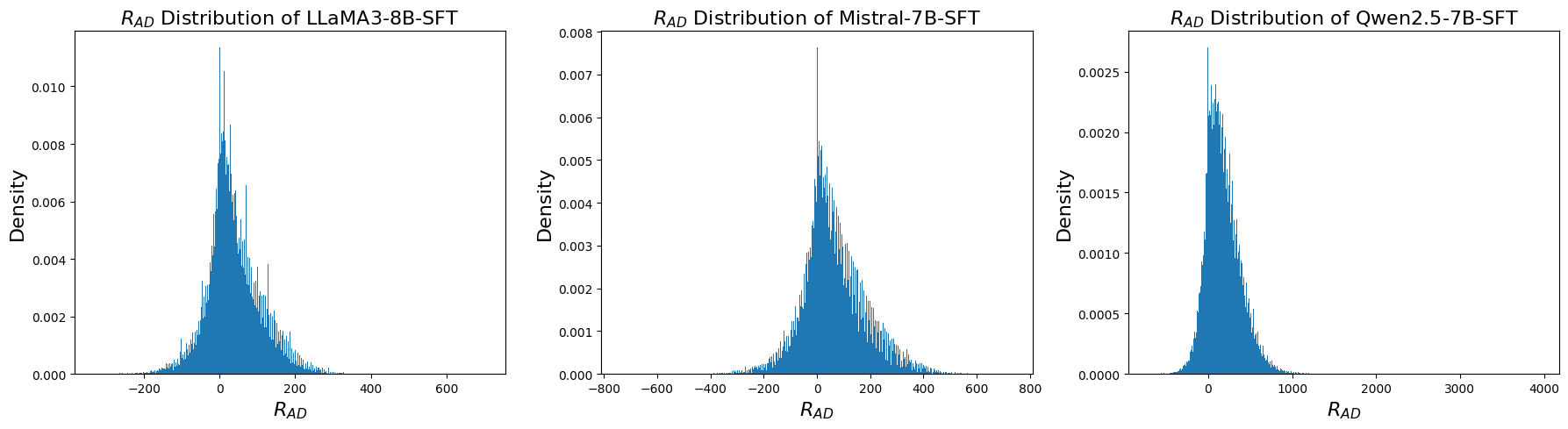}
    \caption{Distribution of $R_{\mathrm{AD}}$ for three SFT models on the UltraFeedback dataset.}
    \label{fig:rad_distribution}
\end{figure*}
\section{Evaluation Settings}
\label{app:evaluation}
\subsection{Evaluation Datasets}
\paragraph{AlpacaEval 2.0}\textit{AlpacaEval 2.0}\footnote{\url{https://github.com/tatsu-lab/alpaca_eval}} is an automated evaluation tool designed to efficiently and cost-effectively assess the performance of instruction-following language models. It is based on the AlpacaFarm dataset and focuses on testing the models’ ability to understand and execute general user instructions. Specifically, we conduct pairwise comparisons on 805 test examples, where the outputs of the DPO-trained model are compared against those of the strong baseline \textit{GPT-4-1106-Preview}. The preferences are judged by the automatic evaluator \textit{deepseek-v3-03-24} (See Appendix~\ref{app:Justification-Evalueator} for the justification of using this as the evaluator). We utilize a fixed decoding temperature (T = 0.9) for all model generation in the experiments. To align with the training phase, we also limit the total length of input and generated tokens to 2048 (max\_length=2048) during inference.
\paragraph{Arena-Hard}\textit{Arena-Hard}\footnote{\url{https://github.com/lmarena/arena-hard-auto}} is a benchmark dataset for evaluating the performance of language models in various tasks, including natural language understanding, reasoning, and generation. It is designed to be challenging and diverse, covering a wide range of topics and domains. We conduct the evaluation on Arena-Hard, using \textit{GPT-4o-mini} as the judge model.
\paragraph{MT-Bench} \textit{MT-Bench}\footnote{\url{https://github.com/lm-sys/FastChat/tree/main/fastchat/llm\_judge\#mt-bench}} is a benchmark framework for evaluating the performance of large language models (LLMs) in multi-turn dialogues, designed to address the limitations of traditional evaluations (such as MMLU and HELM) in open-ended tasks and human preference alignment. We conduct the evaluation on FastChat, using \textit{GPT-4 Turbo} as the judge model.
\subsection{Justification for Using DeepSeek-V3 as the Evalueator}
\label{app:Justification-Evalueator}
Due to the extensive experiments conducted in this paper, the use of gpt4\_turbo in the official \textit{AlpacaEval 2.0} implementation is expensive for us. Therefore, we opted for the more powerful and cost-effective model \textit{deepseek-v3-0324}. We analyzed our annotator using the analyze\_evaluators command in \textit{AlpacaEval 2.0} and compared it with the official annotator in Table~\ref{app:table-Justification-Evalueator}. The results show that our annotator outperforms the official one in human consistency, Spearman correlation, and Pearson correlation, while being significantly more cost-effective.
\begin{table*}[ht]
    \centering
    \caption{Comparison of AlpacaEval 2.0 Annotators. The better results under each metric in the table are highlighted in \textbf{bold}.}
    \resizebox{0.9\textwidth}{!}{
    \begin{tabular}{lcccccc}
        \toprule
        \textbf{Annotator}      & \textbf{Human Agreement} & \textbf{Price} &\textbf{ Spearman Corr.} & \textbf{Pearson Corr.} & \textbf{Bias} & \textbf{Variance} \\ \hline
        DeepSeek-V3    & \textbf{67.27}           & \textbf{0.12}  & \textbf{0.95}           & \textbf{0.87}          & \textbf{32.19} & \textbf{16.45}     \\
        GPT-4 Turbo    & 65.73           & 4.32  & 0.78           & 0.77          & 33.90 & 23.65     \\
        \bottomrule
    \end{tabular}}
    \label{app:table-Justification-Evalueator}
\end{table*}
\section{Details of DPO Training}
\label{app:training_settings}
Following the previous works~\citep{lee2024aligningthousandspreferencesmessage,hu2024openrlhf}, we adopt the \textit{OpenRLHF}~\citep{hu2024openrlhf} framework to perform DPO training, strictly following the standard procedure. All DPO experiments are conducted on 8$\times$ L40 GPUs. Specifically, we use the AdamW optimizer with a cosine learning rate scheduler and a learning rate of 5e-7; a warmup ratio of 10\% is applied at the beginning of training. All models are only trained for 1 epoch over the training set. For the hyper-parameter $\beta$ of DPO, we use a fixed value of $\beta = 0.01$. Input sequences are truncated or padded to a maximum length of 2048 tokens.
\section{Details of Baselines}
\label{app:baselines}
\paragraph{External Reward Margin (EM).}
\label{para:em}
As a commonly used filtering strategy, we select four key evaluation dimensions (helpfulness, instruction following, honesty, and truthfulness) and leverage the powerful Qwen2.5-72B-Instruct model to score the chosen and rejected responses for each preference pair. Unlike the conventional approach of assigning discrete integer scores, we follow the method proposed in \citet{lager}, performing probability-weighted aggregation over score tokens to obtain more fine-grained scores for each dimension. Details of the prompt template can be found in Table~\ref{tbl:dimension_scores}. Specifically, we prompt the model to assign a score from 1 to 9 for each dimension, denoted as $s \in \{1, 2, \dots, 9\}$. The score for each dimension is computed as:
\begin{equation}
\mathrm{Score} = \sum_{s=1}^{9} s \times P(s),
\end{equation}
where $P(s)$ denotes the probability assigned by the model to the score token s, typically obtained via softmax. This formulation allows for a smoother and more fine-grained score. The final score for each sample pair is then computed as the average of the scores across the four dimensions. We then consider the samples with the largest score difference between chosen and rejected as high-quality preference examples for filtering.
\paragraph{PPLGAP.}
To identify high-quality preference pairs, we use PPL Gap as a difficulty-based filtering metric. Given a preference pair $(\mathbf{y}_w, \mathbf{y}_l)$, we define:
\begin{equation}
\text{PPLGap}(\mathbf{y}_w, \mathbf{y}_l) = \text{PPL}(\mathbf{y}_w) - \text{PPL}(\mathbf{y}_l),
\end{equation}
where PPL is computed as:
\begin{equation}
\resizebox{0.4\textwidth}{!}{$
    \text{PPL}(\mathbf{y}) = \exp\left( \frac{1}{|\mathbf{y}|} \sum_{t=1}^{|\mathbf{y}|} -\log P(y_t | y_{<t}) \right).$}
\end{equation}
A larger PPLGap indicates the chosen response is much more likely under the model than the rejected one, suggesting a clearer preference. We select preference pairs with the highest PPLGap values as high-quality data.
\paragraph{Implicit reward margin (IM).}
We use the corresponding SFT model as the reference model and train the policy model on the entire UltraFeedback\_Binarized dataset. Then, we compute the implicit reward margin for all data in this dataset using these three SFT models, as defined in Equation~\ref{eqn:IM}. Samples with the highest implicit reward margins are regarded as high-quality preference examples, as they better reflect a strong alignment between the model and human preferences.
\paragraph{$\mathbf{\text{M}_{AP}}$.}
Following \citet{huang2025larger}, we use the alignment potential metric $M_{AP}$ to assess the quality of preference pairs. It is defined as:
\begin{equation}
\resizebox{0.4\textwidth}{!}{$
M_{AP}(\mathbf{y}_w, \mathbf{y}_l | \mathbf{x}) = \left| M_{ex}(\mathbf{y}_w, \mathbf{y}_l | \mathbf{x}) \right| - \left| M_{im}(\mathbf{y}_w, \mathbf{y}_l | \mathbf{x}) \right|,$}
\end{equation}
where$M_{ex}(\mathbf{y}_w, \mathbf{y}_l | \mathbf{x}) = r(\mathbf{x}, \mathbf{y}_w) - r(\mathbf{x}, \mathbf{y}_l).$
Here, $M_{ex}$ denotes the explicit reward margin given by the reward model, and $M_{im}$ is the implicit reward margin defined in Equation~\ref{eqn:IM}. Intuitively, $M_{AP}$ measures the gap between the target explicit preference and the model's current implicit preference. Therefore, samples with larger $M_{AP}$ are considered to have greater alignment potential, and we rank all preference pairs by $M_{AP}$ and select the top-scoring ones as high-quality training data.
\paragraph{External \& Implicit reward margin (IM\&EM).}
We aggregate the external reward margin $M_{ex}$ and the implicit reward margin $M_{im}$ following the method described in \citet{deng2025moreimprovingllmalignment}. Specifically, we first transform the margin values into margin-guided probabilities through a simple linear transformation:
\begin{equation}
    \mathbb{P}(M) = \frac{\mathrm{clip}(M,M_1,M_2)-M_1}{M_2-M_1},
\end{equation}
where $M \in\{M_{ex},M_{im}\}$, $\mathrm{clip}(M) = min(max(M, M_1), M_2))$ and $(M_1, M_2)$ are tuning parameters. As provided in the original paper, we adopt the optimal settings for $M_1$ and $M_2$. Consequently, we obtain:
\begin{equation}
\resizebox{0.4\textwidth}{!}{$
\begin{aligned}
    &P(\mathbf{y}_w \geq \mathbf{y}_l | M_{\mathrm{ex}}, M_{\mathrm{im}}) \\
    &\quad = \frac{P(M_{\mathrm{ex}}) P(M_{\mathrm{im}})}{P(M_{\mathrm{ex}}) P(M_{\mathrm{im}}) + (1 - P(M_{\mathrm{ex}}))(1 - P(M_{\mathrm{im}}))}.
\end{aligned}$}
\end{equation}
We then select the samples with the highest probabilities obtained above as high-quality preference examples.
\paragraph{R.I.P.}
\citet{rip} proposed that, under the condition of maintaining a positive external reward margin, preference pairs in which the rejected response exceeds a certain length threshold are more likely to be high-quality. Therefore, following the setup in the original paper, we set the external reward margin threshold to 0.126 and select those samples with the longest rejected responses as high-quality preference data.
\paragraph{SDPO.}
\citet{gao2025principleddataselectionalignment} proposed that, by training six reference models to compute validation loss for identifying sample difficulty, then filtering out overly difficult samples that exceed the model's capacity, and conducting preference alignment training only on easy samples within the model's capacity. 
The validation loss $\mathrm{VL(\mathbf{y}_w,\mathbf{y}_l|\mathbf{x})}$ can be computed as:
\begin{equation}
\resizebox{0.4\textwidth}{!}{$
\begin{aligned}
    &\mathrm{VL(\mathbf{y}_w,\mathbf{y}_l|\mathbf{x})} = \\ &-\log \sigma \left( \beta \log \frac{\pi_{\boldsymbol{\theta}}(\mathbf{y}_w | \mathbf{x})}{\pi_{\text{ref}}(\mathbf{y}_w | \mathbf{x})} - \beta \log \frac{\pi_{\boldsymbol{\theta}}(\mathbf{y}_l | \mathbf{x})}{\pi_{\text{ref}}(\mathbf{y}_l | \mathbf{x})}\right).
\end{aligned}$}
\end{equation}
\textbf{In essence, the validation loss $\mathrm{VL(\mathbf{y}_w,\mathbf{y}_l|\mathbf{x})}$ is equivalent to the implicit reward margin. This implies that the data selected by their method can also be filtered using implicit reward margins derived from six different models.} We directly use the top 50\% examples selected from {UltraFeedback\_Binarized} as identified by the paper, which represents the best-performing setting. 

\section{Mathematical Derivations}
\subsection{The Detailed Derivation of Alignment Discrepancy}
\label{app:derivation-of-ad}
The Alignment Discrepancy $ R_{\mathrm{AD}}(\mathbf{y}_w, \mathbf{y}_l | \mathbf{x}) $ is defined based on the difference in implicit reward margins between $\pi_{\boldsymbol{\theta}}^{\text{pos}}$ and $\pi_{\boldsymbol{\theta}}^{\text{inv}}$, with the formula: $R_{\mathrm{AD}}(\mathbf{y}_w, \mathbf{y}_l | \mathbf{x}) = \mathrm{M}^{\mathrm{pos}}_{\mathrm{im}}(\mathbf{y}_w, \mathbf{y}_l | \mathbf{x}) - \mathrm{M}^{\mathrm{inv}}_{\mathrm{im}}(\mathbf{y}_w, \mathbf{y}_l | \mathbf{x}).$ 
Based on the calculation formula for the implicit reward margin as shown in Equation~\ref {eqn:IM}, we can substitute and simplify it to obtain the final form:
\begin{equation}
\resizebox{0.43\textwidth}{!}{$
\begin{aligned}
R_{\mathrm{AD}}&(\mathbf{y}_w, \mathbf{y}_l | \mathbf{x})= \mathrm{M}^{\mathrm{pos}}_{\mathrm{im}}(\mathbf{y}_w, \mathbf{y}_l | \mathbf{x}) - \mathrm{M}^{\mathrm{inv}}_{\mathrm{im}}(\mathbf{y}_w, \mathbf{y}_l | \mathbf{x}) \\
&= \left[ r_{\text{im}}^{\mathrm{pos}}(\mathbf{y}_w | \mathbf{x}) - r_{\text{im}}^{\mathrm{pos}}(\mathbf{y}_l | \mathbf{x}) \right] \\
&\quad - \left[ r_{\text{im}}^{\mathrm{inv}}(\mathbf{y}_w | \mathbf{x}) - r_{\text{im}}^{\mathrm{inv}}(\mathbf{y}_l | \mathbf{x}) \right] \\
&= \left( \log \frac{\pi^{\mathrm{pos}}_{\theta}(\mathbf{y}_w | \mathbf{x})}{\pi_{\text{ref}}(\mathbf{y}_w | \mathbf{x})} - \log \frac{\pi^{\mathrm{pos}}_{\theta}(\mathbf{y}_l | \mathbf{x})}{\pi_{\text{ref}}(\mathbf{y}_l | \mathbf{x})} \right) \\
&\quad - \left( \log \frac{\pi^{\mathrm{inv}}_{\theta}(\mathbf{y}_w | \mathbf{x})}{\pi_{\text{ref}}(\mathbf{y}_w | \mathbf{x})} - \log \frac{\pi^{\mathrm{inv}}_{\theta}(\mathbf{y}_l | \mathbf{x})}{\pi_{\text{ref}}(\mathbf{y}_l | \mathbf{x})} \right) \\
&= \log \frac{\pi^{\mathrm{pos}}_{\theta}(\mathbf{y}_w | \mathbf{x})}{\pi^{\mathrm{pos}}_{\theta}(\mathbf{y}_l | \mathbf{x})} - \log \frac{\pi^{\mathrm{inv}}_{\theta}(\mathbf{y}_w | \mathbf{x})}{\pi^{\mathrm{inv}}_{\theta}(\mathbf{y}_l | \mathbf{x})}.
\end{aligned}$}
\end{equation}
\section{Calculation of GPU Hours for the Methods}
\label{app:cal_gpu_hours}
In our experiments, since both training and inference were consistently conducted on 8×L40 GPUs, it is straightforward to estimate the computational cost of each method. For the four main methods compared in this paper, the estimates are as follows:
\begin{itemize}[leftmargin=1em]
    \item \textbf{EM}: This method requires using the Qwen2.5-72B-Instruct model to score along four dimensions. Deploying this model requires 4 L40 GPUs, and the inference for a single response takes approximately 4 hours. Since each prompt corresponds to two responses (chosen and rejected), the estimated computational cost of EM is: $4 \times 8 \times 4 = 128 \text{ GPUh}$.
    \item \textbf{IM}: This method requires training a forward DPO model on the original dataset, which takes about 8 GPUs × 4h = 32 GPUh. Then, computing the implicit reward margin takes about 11 GPUh. Thus, the total cost of IM is: $32 + 11 = 43 \text{ GPUh}$.
\item \textbf{\mname}: This method requires training a forward DPO model on the original dataset and a reverse DPO model on the inverted dataset, taking about 8 × 4 × 2 = 64 GPUh in total. In addition, computing the forward and reverse implicit reward margins and the NLL of the original model takes about 11 + 5.5 = 16.5 GPUh. Therefore, the total cost of \mname is:
$64 + 16.5 = 80.5 \text{ GPUh}$.
\item \textbf{SDPO}: This method requires training 6 reference models on half of the original dataset, which costs about 6 × 8 × 2 = 96 GPUh. Then, each model requires implicit reward margin computation, costing about 6 × 11 = 66 GPUh. Thus, the total cost of SDPO is:
$96 + 66 = 162 \text{ GPUh}$.
\end{itemize}
\section{Prompt Template}
\label{prompt}
\begin{figure*}
    \centering
    \includegraphics[width=1\linewidth]{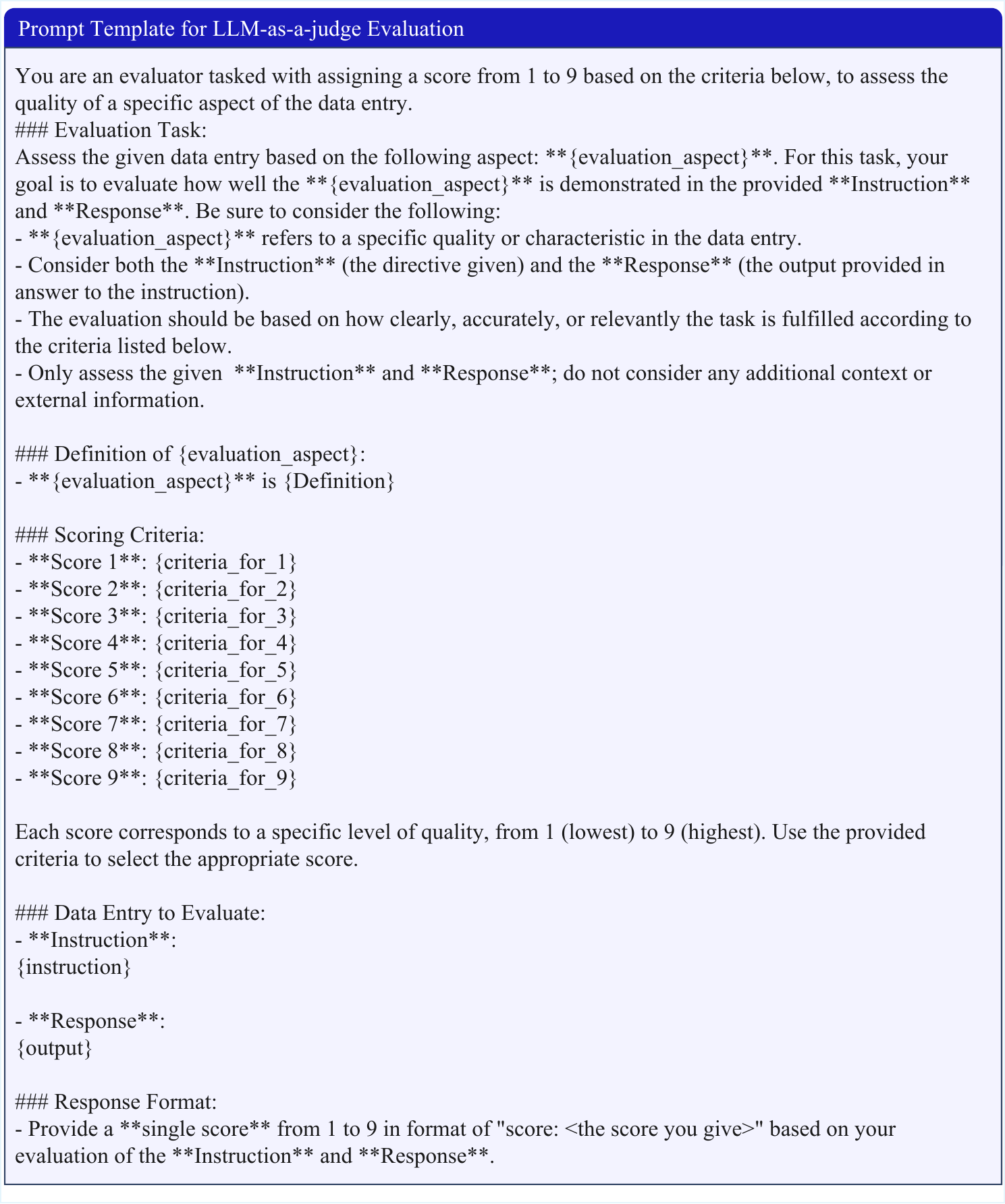}
    \caption{\textbf{Prompt Template for Scoring Preference Pairs Using Qwen2.5-72B-Instruct}.}
    \label{fig:template}
\end{figure*}

In this section, we present all the prompt templates used in this paper. To evaluate preference pairs effectively, we design a structured prompt template that guides the Qwen2.5-72B-Instruct model in assigning scores across multiple evaluation dimensions. The specific design of our template is illustrated in Figure~\ref{fig:template}, which provides a clear example of how questions, responses, and scoring instructions are presented to the model. Specifically, we focus on four key dimensions: helpfulness, instruction following, honesty, and truthfulness. The detailed definitions of different dimensions and their specific scoring standards are presented in Table~\ref{tbl:dimension_scores}.

\newpage
\onecolumn
 
\begin{longtblr}[
    caption = {Definitions of different dimensions and their specific scoring standards, with each dimension scored on a scale ranging from 1 to 9, each score corresponds to a specific standard.},
  label = {tbl:dimension_scores}
]{
  width = \textwidth,       % span full text width (both columns)
  colspec = {Q[150]Q[812]},
  row{4} = {t},
  column{1} = {c},
  cell{2}{1} = {r=9}{},
  cell{12}{1} = {r=9}{},
  cell{22}{1} = {r=9}{},
  cell{32}{1} = {r=9}{},
  cell{42}{1} = {r=9}{},
  cell{52}{1} = {r=9}{},
  cell{62}{1} = {r=9}{},
  vlines,
  hline{1,71} = {-}{0.12em},
  hline{2,11-12,21-22,31-32,41-42,51-52,61-62} = {-}{},
  hline{3-10,13-20,23-30,33-40,43-50,53-60,63-70} = {2}{},
}
\textbf{Dimension-1 \& Definition} & \textbf{Answer Accuracy:} Evaluate whether the response accurately addresses the instruction and completely fulfills the task.\\
\textbf{Scoring Standards} & 1: Completely incorrect, irrelevant to the instruction.\\
 & 2: Partially correct, major omissions or errors in fulfilling the instruction.\\
 & 3: Contains significant errors, unable to fully address the core task.\\
 & 4: Partially correct, missing key details or addressing the wrong aspect of the instruction.\\
 & 5: Mostly accurate, but contains some errors or omissions.\\
 & 6: Mostly correct, though missing small details or has minor inaccuracies.\\
 & 7: Largely accurate and complete, but may lack small details or have minimal errors.\\
 & 8: Fully accurate, completely addresses the instruction, minimal flaws.\\
 & 9: Perfectly accurate, fully aligns with the instruction, no omissions.\\
\textbf{Dimension-2 \& Definition} & \textbf{Helpfulness:} Assess whether the response provides valuable and actionable information that directly aids the user in achieving their objective.\\
\textbf{Scoring Standards} & 1: Response is entirely unhelpful and irrelevant, offering no value.\\
 & 2: Response is mostly unhelpful, with very little relevance to the user's needs.\\
 & 3: Provides limited help, but the response is vague, incomplete, or only partially relevant.\\
 & 4: Response is somewhat helpful, but important details or steps are missing.\\
 & 5: Moderately helpful, addressing the main query but lacking depth or precision.\\
 & 6: Mostly helpful and clear, with minor gaps or room for further detail.\\
 & 7: Helpful and mostly comprehensive, fulfilling the majority of the user's request.\\
 & 8: Highly helpful and clear, with only minimal areas for potential improvement.\\
 & 9: Exceptionally helpful, fully addressing the user's needs in a complete and actionable manner.\\
\textbf{Dimension-3 \& Definition} & \textbf{Honesty:} Evaluate whether the response is truthful and avoids misleading or fabricating information.\\
\textbf{Scoring Standards} & 1: Response is entirely dishonest, fabricated, or intentionally misleading.\\
 & 2: Response includes significant falsehoods or misleading information.\\
 & 3: Some information is correct, but key elements are false or deceptive.\\
 & 4: Mostly truthful, but contains noticeable inaccuracies or vague claims.\\
 & 5: Generally honest, though minor factual errors or omissions exist.\\
 & 6: Honest and clear, with very few small inaccuracies or overly cautious phrasing.\\
 & 7: Mostly accurate and honest, with no significant misleading elements.\\
 & 8: Entirely honest and accurate, only minor potential for clarification needed.\\
 & 9: Flawlessly honest, all information is accurate and presented transparently.\\
\textbf{Dimension-4 \& Definition} & \textbf{Instruction Following:} Determine if the response adheres precisely to the instructions provided, fulfilling the request as intended.\\
\textbf{Scoring Standards} & 1: Completely disregards the instruction, with no relation to the request.\\
 & 2: Fails to follow the instruction significantly, with only minor relevant elements.\\
 & 3: Partially follows the instruction, but with major omissions or errors.\\
 & 4: Somewhat follows the instruction, though important aspects are overlooked.\\
 & 5: Follows the instruction moderately well, with noticeable gaps or misinterpretations.\\
 & 6: Mostly adheres to the instruction, with minor deviations or missed nuances.\\
 & 7: Adheres to the instruction well, with only slight areas for improvement.\\
 & 8: Accurately follows the instruction, with minimal need for refinement.\\
 & 9: Perfectly adheres to the instruction, fulfilling every aspect flawlessly.\\
\textbf{Dimension-5 \& Definition} & \textbf{Truthfulness:} Assess whether the response is factually accurate and based on verified knowledge or reasoning.\\
\textbf{Scoring Standards} & 1: Response is completely false, with no factual basis or accuracy.\\
 & 2: Response is mostly false, with very few correct facts.\\
 & 3: Response has a mix of true and false information, with major inaccuracies.\\
 & 4: Response is somewhat accurate but includes noticeable factual errors.\\
 & 5: Response is generally accurate but contains some minor factual inaccuracies.\\
 & 6: Mostly truthful and fact-based, with minimal errors or ambiguities.\\
 & 7: Highly accurate and truthful, with no significant errors or misleading information.\\
 & 8: Entirely truthful and accurate, with only trivial areas for clarification or nuance.\\
 & 9: Flawlessly truthful, presenting facts with utmost accuracy and precision.\\
\end{longtblr}

\end{document}